\documentclass{article}

 \usepackage[main, final]{neurips_2026}

\usepackage{xcolor}
\definecolor{lightblue}{rgb}{0.12,0.49,0.85}

\usepackage{graphicx}
\usepackage{amsmath} 

\usepackage[utf8]{inputenc} 
\usepackage[T1]{fontenc}    
\usepackage[pagebackref,breaklinks,colorlinks,citecolor=lightblue]{hyperref}
\usepackage{url}            
\usepackage{booktabs}       
\usepackage{amsfonts}       
\usepackage{nicefrac}       
\usepackage{microtype}      
\usepackage{xcolor}         

\usepackage{orcidlink}
\usepackage{multirow}
\usepackage{wrapfig}
\usepackage{cleveref}

\title{GenCOPE: Syn2Real Generalized Category-Level Object Pose Estimation for Robotic Picking}

\author{Jian Liu$^{1}$, Wei Sun$^{2}$, Zhenqi Dai$^{2}$, Hui Yang$^{2}$, Jian Xiao$^{2}$, Nicu Sebe$^{3}$, Na Zhao$^{1,\dagger}$
\\[0.2cm]
$^{1}$ IMPL Lab, SUTD~~~
$^{2}$ NERC-RVC, Hunan University~~~
$^{3}$ University of Trento
\\[0.2cm]
\textbf{Project page:} \href{https://paperreview99.github.io/GenCOPE/}{CNJianLiu/GenCOPE}
}

\begin{document}

\maketitle

\begingroup
  \renewcommand{\thefootnote}{\fnsymbol{footnote}}
  \footnotetext[2]{Corresponding author.}
\endgroup

\vspace{-0.2in}
\begin{figure}[h]
\begin{center}
\includegraphics[width=1\textwidth]{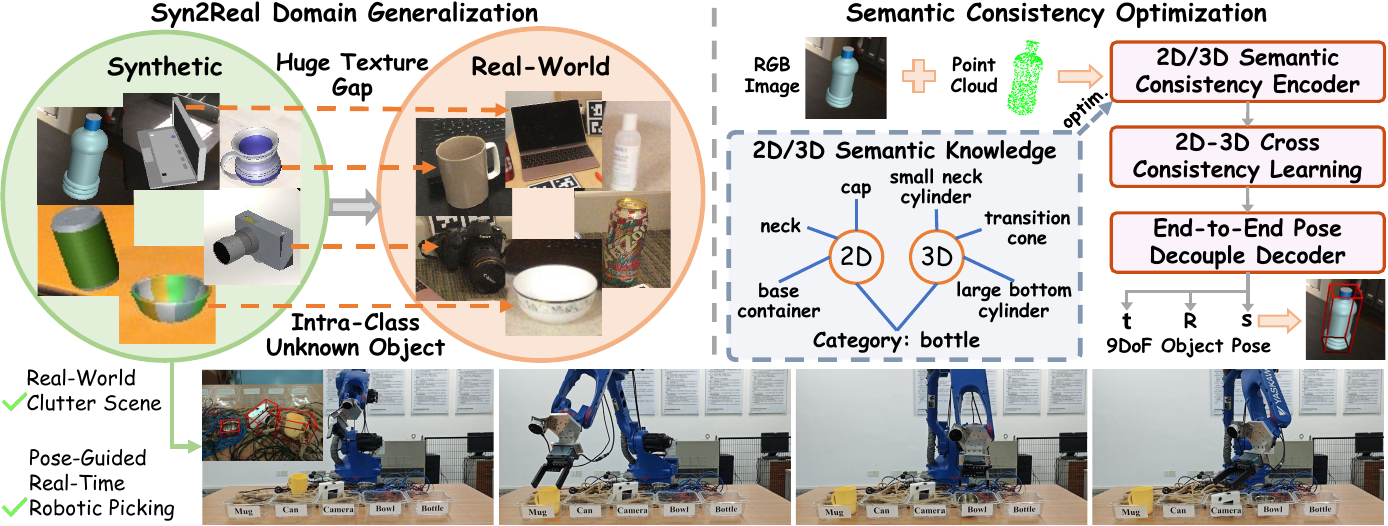}
\end{center}
\vspace{-0.1in}
\caption{GenCOPE for real-world robotic picking. \emph{Left}: synthetic-to-real (Syn2Real) domain-generalized task setting, which aims to achieve generalization between intra-class unknown objects from synthetic to real-world domains. \emph{Right}: 2D/3D semantic consistency optimization for end-to-end pose regression, where the dashed box indicates that it is used for optimization only during training.}
\label{Fig1}
\end{figure}

\begin{abstract}
Category-level object pose estimation (COPE), capable of generalizing to intra-class unknown objects, has become a core technique for robotic 3D scene understanding. However, existing COPE methods still require labor-intensive recollection of real-world training data for novel object categories, which limits their scalability in practical applications. This paper aims to achieve synthetic-to-real (Syn2Real) generalized COPE, where a model is trained solely on rendered synthetic data and directly generalized to real-world deployments. The central challenge lies in the significant domain gap between synthetic and real-world data, particularly in texture appearance. To address this, we aim to enhance domain generalization by learning domain-invariant representations that capture semantic commonalities among objects within the same category. We introduce 2D and 3D semantic consistency constraints to reduce the sensitivity of feature encoders to domain-specific features. In addition, we propose an end-to-end pose regression framework that performs 2D-3D cross consistency learning, leveraging dense cross-modality fusion to further refine pose estimation. Since simplicity and effectiveness are essential for real-world robotic deployment, our model operates exclusively on global features, yielding a highly lightweight and efficient architecture. Extensive experiments on the REAL275 and Wild6D benchmarks, as well as real-world robotic manipulation scenes, show superior Syn2Real generalization performance of our paradigm.
\end{abstract}

\section{Introduction}
\vspace{-0.25em}
Nine-degrees-of-freedom (9DoF) object pose estimation, which aims to estimate the 3DoF translation, 3DoF rotation, and 3DoF size of an object in the camera coordinate system \cite{liu2026survey, NOCS, query6dof, GeoReF}, is a fundamental problem in robotic 3D scene understanding and manipulation \cite{PhoCaL, HouseCat6D, LanCOPE, zhang2024omni6dpose, huang2025givepose, liu2026sinref6d}. Early studies primarily focused on instance-level pose estimation \cite{DenseFusion, PVN3D, FFB6D, GDR-Net, yang2026occlusion}, where models are restricted to objects seen during training. Alternatively, category-level object pose estimation (COPE) has emerged as a promising direction \cite{Synthesis, CR-Net, OLD-Net, BundleTrack, CATRE, AG-Pose, monodiff9d}, as it enables generalization to intra-class unknown objects without relying on explicit prior information such as CAD models or reference views.

Despite the impressive generalization capability of category-level approaches across unknown instances within the same category, existing methods still require labor-intensive collection and annotation of real-world training data for each novel object category \cite{mkpose, yang2026genhoi, ren2026compose, lin2026mempose}. This requirement significantly hinders scalability and limits their practicality in large-scale robotic applications \cite{zhao2020sess, zhao2021few, yang2025rgb, geng2026disentangled}. In contrast, recent advances in rendering technology have made large-scale synthetic data readily available, with accurate pose annotations obtained at negligible cost. Motivated by this observation, this paper aims to achieve Syn2Real domain-generalized COPE, referred to as GenCOPE, where models trained purely on synthetic data can be directly deployed in real world.

The central challenge in GenCOPE lies in the significant domain gap between synthetic and real-world data. Through an analysis of the rendered synthetic data used for object pose estimation \cite{NOCS}, we observe that the most pronounced discrepancy arises from texture appearance, as illustrated in Fig. \ref{Fig1}. 
However, existing methods primarily rely on feature extractors \cite{pspnet, ViT} that emphasize fine-grained image features \cite{DPDN, IST-Net, miao2026langmap}, which inherently encode substantial texture information while failing to explicitly capture domain-invariant properties shared between synthetic and real-world data. Consequently, the learned representations become overly sensitive to texture variations, making them suboptimal for robust Syn2Real domain-generalized pose estimation.

To address the aforementioned challenge, we argue that although intra-class unknown objects exhibit substantial texture discrepancies between synthetic and real-world domains, their 2D and 3D semantic structures remain largely consistent. For instance, at the 2D level, objects such as \emph{bottles} typically consist of semantic parts including \emph{a cap, a neck, and a base container}. Similarly, at the 3D level, their geometric composition can be abstracted as \emph{a small neck cylinder, a transitional cone, and a large bottom cylinder}. Based on this observation, we posit that explicitly incorporating semantic knowledge as constraints during 2D and 3D feature encoding can encourage encoders to focus on domain-invariant semantic cues, thereby suppressing the adverse effects of domain-specific features. An overview of the proposed GenCOPE is illustrated in Fig. \ref{Fig1}.

Given a segmented RGB image and its corresponding point cloud (obtained via depth map back-projection) as inputs, we first perform feature embedding using the proposed 2D/3D semantic consistency encoders, followed by a 2D-3D cross consistency learning module and an end-to-end pose decouple decoder. During training, semantic consistency is enforced by leveraging semantic supervision derived from a pretrained large language model \cite{team2023gemini} and the CLIP text encoder \cite{clip}, which aligns the 2D and 3D feature representations in a shared semantic space. The resulting semantically consistent features are then fused through a 2D-3D cross transformer-based module to perform cross-consistency learning \cite{yang2022deepinteraction}. Finally, the fused features are concatenated and fed into a pose decouple decoder to regress the full 9DoF object pose.
To prevent feature extractors from overemphasizing fine-grained texture-dependent features while maintaining computational efficiency, we adopt the most lightweight backbones based on ResNet18 \cite{ResNet} and PointNet \cite{PointNet} as the 2D and 3D encoders, respectively, and operate exclusively on global features throughout the pipeline.
This design is deliberately adopted to jointly enhance domain generalization robustness while meeting the real-time requirements of practical robotic manipulation scenarios.

Overall, our main contributions are as follows: (1) We propose a simple yet effective framework for Syn2Real domain-generalized COPE, explicitly designed for real-world robotic deployment. The framework operates purely on global features from most lightweight backbones, avoiding dense local texture correspondences, making it highly efficient under resource-constrained settings. (2) We introduce semantic consistency encoders in both 2D and 3D spaces to learn semantic-invariant features shared across intra-class objects, alleviating the severe gap between synthetic and real-world domains. (3) We propose an end-to-end pose estimation pipeline based on 2D-3D cross consistency learning, featuring a cross-modal fusion network and a pose decouple decoder, enabling mutual reinforcement across modalities for more accurate and robust pose estimation.

\section{Related Work}
\noindent\textbf{Fully-Supervised COPE Methods.}
NOCS \cite{NOCS} introduces a normalized object coordinate space formulation and establishes two widely adopted benchmarks that advance the research of COPE methods. To address substantial shape variability among intra-class objects, SPD \cite{SPD} first leverages category-level CAD model libraries to extract shape priors offline and subsequently reconstructs NOCS representations through shape prior deformation for pose estimation. This paradigm has inspired a series of follow-up approaches that explicitly incorporate shape priors \cite{SGPA, centersnap, RBP-Pose, SSP-Pose}.
Despite their effectiveness in modeling intra-class shape variability, prior-based approaches rely on manually constructed category-level CAD model libraries, which incur substantial human effort.

The limitation aforementioned has motivated the development of prior-free alternatives \cite{DualPoseNet, FS-Net, GPV-Pose, IST-Net, AG-Pose, GenPose, clipose, SpotPose, cleanpose}.
FS-Net \cite{FS-Net} proposes a 3D graph convolution-based network and performs decoupled pose regression. GPV-Pose \cite{GPV-Pose} enhances the learning of pose-sensitive features via geometry-guided point-wise voting.
Beyond these depth-only approaches, several RGB-D methods \cite{IST-Net, VI-Net} jointly leverage appearance and geometric cues.
SecondPose \cite{SecondPose} introduces DINOv2 \cite{DINOv2} to provide SE(3)-consistent semantic features for pose estimation.
GCE-Pose \cite{GCE-Pose} enhances COPE under partial visibility by reconstructing category-level global context via semantic shape reconstruction and fusing it with local RGB-D observations. CleanPose \cite{cleanpose} mitigates data bias by integrating causal inference with front-door adjustment and residual-based knowledge distillation from 3D foundation models.
Although these fully-supervised methods achieve competitive performance, their dependence on extensive real-world data annotation makes them difficult to scale to novel object categories.

\noindent\textbf{Label-Efficient COPE Methods.}
To alleviate the scalability bottleneck caused by extensive real-world data annotation, recent studies have explored label-efficient COPE \cite{CPPF, SAR-Net, Wild6D, SSC6D, DPDN, UDA-COPE, TTA-COPE, MH6D, diffusionnocs, Diff9D}. CPPF \cite{CPPF} introduces a category-level point pair feature voting formulation that enables pose estimation in the wild without dense supervision. SAR-Net \cite{SAR-Net} investigates object shape alignment by directly matching the observed point cloud with shape prior, reducing the reliance on real-world training data. DPDN \cite{DPDN} designs a domain-generalized learning framework based on parallel deep prior deformation. UDA-COPE \cite{UDA-COPE} leverages a teacher-student self-supervised learning scheme to achieve domain adaptation. Further, TTA-COPE \cite{TTA-COPE} proposes a test-time adaptation strategy that first learns from labeled synthetic data and then adapts the pretrained model online to real-world data during inference. More recently, DiffusionNOCS \cite{diffusionnocs} adopts diffusion models to predict dense canonical maps for pose estimation, demonstrating strong robustness to symmetry and effective Syn2Real generalization. Diff9D \cite{Diff9D} directly formulates COPE as a diffusion-based generative process, enabling Syn2Real generalization without requiring 3D shape priors.

Although existing methods perform well, they do not explicitly model the consistent properties shared between synthetic and real-world data. In contrast, this paper analyzes the domain-invariant characteristics across synthetic and real domains and explicitly incorporates semantic knowledge as constraints during both 2D and 3D feature encoding. This design encourages encoders to focus on domain-invariant semantic cues, alleviating the adverse impact of domain-specific information.

\section{Workflow of GenCOPE}
\textbf{Task definition}: GenCOPE aims to predict the 6DoF object pose (consisting of 3DoF translation and 3DoF rotation) and 3DoF object size (axes lengths of oriented bounding box) of intra-class unknown objects in real world using the model trained only on synthetic dataset.

\subsection{Overview}
GenCOPE takes a single RGB-D image as input. The target object is first segmented using Mask R-CNN \cite{MaskR-CNN} to obtain an object-centric RGB image $I$ \cite{miao2023spectrum, miao2024referring}. The corresponding segmentation mask is then applied to crop the depth map, which is back-projected to generate a point cloud $P$ with 1024 sampled points. This preprocessing follows the standard pipeline adopted by most prior works \cite{SGPA, cleanpose, Diff9D}. Subsequently, $I$ and $P$ are processed by dedicated 2D and 3D encoders equipped with semantic consistency optimization (\emph{applied only during training}) to extract global semantic consistency features $F_{C\text{-}2D}$ and $F_{C\text{-}3D}$. The resulting features are then fused by a proposed 2D-3D cross transformer network, which explicitly models cross-modal consistency. Finally, the fused 2D-3D consistency-aware global features are concatenated and fed into a decoupled decoder to regress the 3DoF translation $\mathbf{t}$, 3DoF rotation $\mathbf{R}$, and 3DoF size $\mathbf{s}$, respectively. An overview of the proposed framework is illustrated in Fig. \ref{Fig2}.

\subsection{2D and 3D Semantic Consistency Encoders}
\begin{figure*}[t!]
\centering
\includegraphics[width=\textwidth]{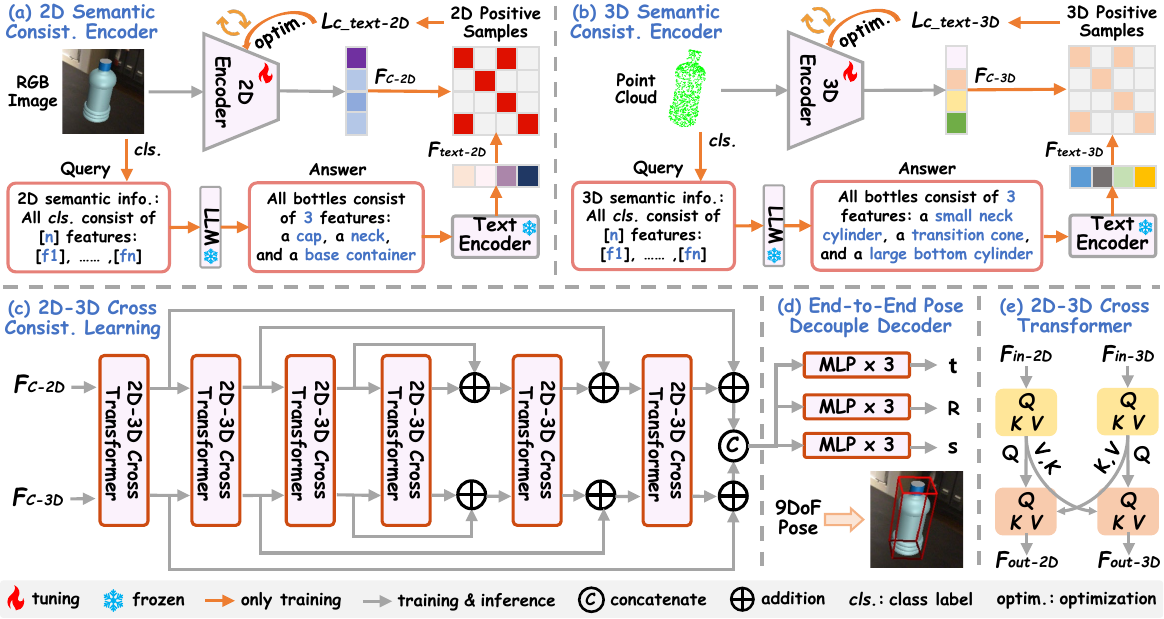}
\caption{Overall workflow of GenCOPE. (a) and (b) illustrate the proposed 2D and 3D semantic consistency encoders, which impose training-time constraints to emphasize intra-class semantic invariances. (c) presents the proposed 2D-3D cross consistency learning module that fuses two-modal features, where the 2D-3D cross transformer architecture is detailed in (e). (d) denotes the end-to-end pose decouple decoder, which regresses the full 9DoF object pose.}
\label{Fig2}
\end{figure*}

To introduce intra-class semantic commonalities into RGB image and point cloud encoders in a simple yet effective manner, we leverage semantic optimization generated by the large language model (LLM) Gemini-3 \cite{team2023gemini}, as illustrated in (a) and (b) of Fig. \ref{Fig2}. Specifically, we define a semantic query template: \emph{Output the common 2D/3D semantic information of a class of objects enclosed in parentheses in the following format: ``All} \emph{cls.} \emph{consist of} [\emph{n}] \emph{features:} [\emph{f$_1$}], \ldots, [\emph{f$_n$}]\emph{''}, where \emph{cls.} denotes the segmented object class, and the number of features [\emph{n}] as well as the semantic descriptors [\emph{f$_1$}]-[\emph{f$_n$}] are inferred by the LLM. To enhance the diversity of semantic descriptions and mitigate the impact of uncertainty, we prompt the LLM to generate \textbf{ten distinct} semantic descriptions for each object class and randomly select one description for each training sample. Since this semantic consistency optimization is applied only in the training phase, the \textbf{ground-truth class label} \emph{cls.} for each image is directly available. For instance, when \emph{cls.} corresponds to \emph{bottle}, a representative 2D semantic description is \emph{``All bottles consist of three features: a cap, a neck, and a base container''}, while the corresponding 3D semantic description is \emph{``All bottles consist of three features: a small neck cylinder, a transition cone, and a large bottom cylinder''}. Additional LLM-generated semantic examples for different object categories are provided in the Appendix \ref{Additional Semantic Samples Generated by LLM}.

The generated 2D and 3D semantic descriptions are encoded using the CLIP pretrained text encoder \cite{clip}, producing semantic feature embeddings $F_{text\text{-}2D} \in \mathbb{R}^{b \times 512}$ and $F_{text\text{-}3D} \in \mathbb{R}^{b \times 512}$, where $b$ denotes the batch size. These semantic embeddings are then employed to optimize the global features encoding processes of the 2D and 3D encoders \cite{ResNet, PointNet}, respectively, resulting in semantically consistent representations $F_{C\text{-}2D} \in \mathbb{R}^{b \times 512}$ and $F_{C\text{-}3D} \in \mathbb{R}^{b \times 512}$. By explicitly encouraging the encoders to attend to intra-class semantic invariances, the proposed optimization suppresses sensitivity to discrepancies between synthetic and real-world domains, improving domain generalization ability.

We consider a batch consisting of $b$ images and $b$ text descriptions, and let $F_{C\text{-}2D}^i$ and $F_{text\text{-}2D}^i$ denote the 2D feature of the $i$-th image and the semantic feature of the $i$-th text, respectively. Since a training batch may contain multiple positive image-text pairs sharing the same \emph{cls.} label, we define $\mathcal{P}(i)$ (positive sample) as the set of indices of all text samples in the batch whose \emph{cls.} labels are identical to that of $F_{C\text{-}2D}^i$, and $\mathcal{A}(i)$ as the index set of all text samples in $F_{text\text{-}2D}$. Based on these definitions, the 2D semantic consistency loss is formulated as:
\begin{equation}
\mathcal{L}_{C\_text\text{-}2D} = \frac{1}{b} \sum_{i=1}^{b} \mathcal{L}_{i},
\end{equation}
where the loss for each sample $i$, denoted as $\mathcal{L}_{i}$, is defined as:
\begin{equation}\label{equation2}
\mathcal{L}_i
=
-\frac{1}{|\mathcal{P}(i)|}
\sum_{p \in \mathcal{P}(i)}\log
\frac{
\exp\left( z_i \cdot z_p / \tau \right)
}{
\sum\limits_{j \in \mathcal{A}(i)}
\exp\left( z_i \cdot z_j / \tau \right)
},
\end{equation}
where $z_i$ denote the normalized feature vector of the $i$-th sample from $F_{C\text{-}2D}$, $z_p$ and $z_j$ represent normalized feature vectors from $F_{text\text{-}2D}$, $|\mathcal{P}(i)|$ is the cardinality of the positive sample set, and $\tau$ is a scalar temperature parameter \cite{khosla2020supervised}.

Similarly, we can compute the 3D semantic consistency loss $\mathcal{L}_{C\_text\text{-}3D}$ following the same formulation, by replacing $F_{C\text{-}2D}$ and $F_{text\text{-}2D}$ with $F_{C\text{-}3D}$ and $F_{text\text{-}3D}$ in the corresponding batch. By incorporating $\mathcal{L}_{C{\_}text\text{-}2D}$ and $\mathcal{L}_{C{\_}text\text{-}3D}$ during training, we can enforce semantic consistency supervision of the 2D and 3D encoders, discouraging over-reliance on domain-specific information.

\subsection{2D-3D Cross Consistency Learning for Pose Regression}
To effectively align semantic representations across modalities and enable robust pose estimation, we introduce a 2D-3D cross consistency learning framework coupled with an end-to-end pose decouple decoder. As illustrated in (c) and (d) of Fig. \ref{Fig2}, the proposed design jointly fuses semantically consistent features $F_{C\text{-}2D}$ and $F_{C\text{-}3D}$ via stacked cross transformers and decodes the object pose in a fully decoupled end-to-end manner. The details of some modules are provided in the Appendix \ref{Details of Some Modules}.

\subsubsection{2D-3D Cross Consistency Learning}
We construct a stack of $2N$ 2D-3D cross transformer blocks to progressively model cross-modal interactions at multiple semantic levels. Each cross transformer block follows the architecture shown in (e) of Fig. \ref{Fig2}, consisting of a self-attention module and a cross-attention module. For the $k$-th block, self-attention is first applied independently to each modality to capture intra-modal dependencies:
\begin{equation}
\vspace{0.1em}
\begin{aligned}
\tilde{F}_{C\text{-}2D}^{(k)} &= \mathrm{SelfAttn}\!\left(F_{C\text{-}2D}^{(k-1)}\right),\;
\tilde{F}_{C\text{-}3D}^{(k)} = \mathrm{SelfAttn}\!\left(F_{C\text{-}3D}^{(k-1)}\right),
\end{aligned}
\vspace{0.1em}
\end{equation}
where the multi-head attention \cite{ViT} is formulated as:
\begin{equation}\label{equation4}
\mathrm{Attn}\left( \cdot \right) = \mathrm{Concat}\left( {Softmax\left( {{Q}{K{^ \top}}/\sqrt d } \right){V}} \right),
\end{equation}
where Q, K, and V denote the input query, key, and value, respectively. $\top$ represents matrix transpose. $d$ is the feature dimension of each attention head. $\mathrm{Concat}(\cdot)$ denotes concatenate the features of all heads. Then, cross attention is employed to enable inter-modal feature exchange, where queries originate from the same modality and keys/values from the opposite modality:
\begin{equation}
\begin{aligned}
\hat{F}_{C\text{-}2D}^{(k)} &= \mathrm{CrossAttn}\!\left(
Q=\tilde{F}_{C\text{-}2D}^{(k)},\;
K=\tilde{F}_{C\text{-}3D}^{(k)},\;
V=\tilde{F}_{C\text{-}3D}^{(k)}
\right), \\
\hat{F}_{C\text{-}3D}^{(k)} &= \mathrm{CrossAttn}\!\left(
Q=\tilde{F}_{C\text{-}3D}^{(k)},\;
K=\tilde{F}_{C\text{-}2D}^{(k)},\;
V=\tilde{F}_{C\text{-}2D}^{(k)}
\right).
\end{aligned}
\end{equation}

To stably aggregate semantic information across different levels and encourage consistent cross-modal alignment, we introduce a symmetric residual fusion strategy, as shown in (c) of Fig. \ref{Fig2}. The outputs of cross transformer blocks are paired symmetrically from outer to inner layers, which explicitly combines shallow and deep semantic representations \cite{PVNet}.

\subsubsection{End-to-End Pose Decouple Decoder}
The fused features from the outermost symmetric pair are concatenated to form the final joint representation $F_{\text{out}}$.
As shown in (d) of Fig. \ref{Fig2}, $F_{\text{out}}$ is fed into an end-to-end pose decouple decoder composed of multiple multi-layer perceptrons (MLPs), which independently regress translation $\mathbf{t}$, rotation $\mathbf{R}$, and size $\mathbf{s}$, yielding full 9DoF object pose:
\begin{equation}
\mathbf{t},\, \mathbf{R},\, \mathbf{s} \;=\; \mathrm{MLPs}(F_{\text{out}}),\ \mathrm{MLPs}(F_{\text{out}}),\ \mathrm{MLPs}(F_{\text{out}}).
\end{equation}
We follow most previous work \cite{IST-Net, DPDN, GenPose, Diff9D} in using translation, rotation, and size matrices to represent the 9DoF object pose.

\subsection{Training and Inference Protocol}
We train the proposed GenCOPE framework in an end-to-end manner, jointly optimizing semantic consistency learning and pose regression. The overall training objective consists of two components: semantic consistency losses that supervise the feature encoders, and a pose regression loss that guides the decoupled pose decoder. Specifically, the total training loss is defined as:
\begin{equation}
\mathcal{L}_{\text{total}} = \mathcal{L}_{C\_text\text{-}2D} + \mathcal{L}_{C\_text\text{-}3D} + \mathcal{L}_{\text{pose}},
\end{equation}
where the first two items enforce semantic alignment between visual features and text-derived semantic priors within each object category, and $\mathcal{L}_{\text{pose}}$ supervises the regression of the object pose. $\mathcal{L}_{\text{pose}}$ is computed as the sum of the L2 norms of translation, rotation, and size prediction errors \cite{IST-Net, Diff9D}, which supports the pose-decoupled decoder by supervising each pose component independently.

\emph{During training}, the semantic consistency losses $\mathcal{L}_{C\_text\text{-}2D}$ and $\mathcal{L}_{C\_text\text{-}3D}$ are applied to constrain the 2D and 3D encoders to focus on intra-class semantic invariants, thereby reducing reliance on domain-specific cues and improving robustness to domain shifts between synthetic and real-world data. The pose loss $\mathcal{L}_{\text{pose}}$ is jointly optimized to ensure accurate pose estimation. \emph{At inference time}, the semantic consistency supervision parts are removed. Given a single RGB-D image, the trained 2D and 3D encoders, 2D-3D cross consistency learning module, and pose decouple decoder are executed in a single forward pass to directly predict the full 9DoF object pose, \textbf{without requiring any textual input} or additional optimization. This design enables efficient and scalable deployment in real-world robotic applications.

\section{Experiments}
\textbf{Datasets}
We follow previous COPE methods \cite{Wild6D, IST-Net, Diff9D} to utilize the CAMERA25 \cite{NOCS}, REAL275 \cite{NOCS}, and Wild6D \cite{Wild6D} datasets for evaluation. Specifically, we use the large synthetic CAMERA25 for training, which includes 1085 instances distributed in $ \sim $250K synthetic images. During inference, we evaluate on real-world REAL275 and Wild6D datasets and self-collected robotic picking scenes.

\noindent\textbf{Implementation Details}
The input RGB-D images are captured at a resolution of $640 \times 480$. For preprocessing, instance segmentation is first performed using Mask R-CNN \cite{MaskR-CNN}. The resulting object mask is applied to both the RGB and depth images, after which the RGB crop is resized to $192 \times 192$ to reduce computational overhead. The temperature parameter $\tau$ in Eq. \ref{equation2} is set to 0.07 as the previous contrastive learning method \cite{khosla2020supervised}, the feature dimension $d$ in Eq. \ref{equation4} is fixed to 32, and the batch size $b$ is set to 48. Model optimization adopts a cyclic learning rate schedule, where the learning rate oscillates between $1\times10^{-6}$ and $1\times10^{-4}$ using CyclicLR \cite{CyclicLR}, with a step size of 20K iterations per cycle. Our model is trained for a total of 100 cycles. All experiments on public benchmarks are conducted on a single NVIDIA RTX 3090 GPU with 24 GB memory. Real-world experiments are deployed on a physical robotic platform equipped with an NVIDIA GTX 1660 GPU with 6 GB memory and an Intel RealSense L515 RGB-D camera.

\noindent\textbf{Evaluation Metrics}
Following common practice in prior work \cite{SGPA, CATRE, SecondPose, cleanpose, Diff9D}, we evaluate performance using the $n^\circ m\mathrm{cm}$ and 3D Intersection-over-Union ($IoU_{3D}$) metrics. The $n^\circ m\mathrm{cm}$ criterion measures pose accuracy by jointly thresholding rotation and translation errors, where a prediction is deemed correct if the rotation and translation errors fall below $n^\circ$ and $m\mathrm{cm}$, respectively. We report results under $5^\circ 2{\rm{cm}}$, $5^\circ 5{\rm{cm}}$, $10^\circ 2{\rm{cm}}$, and $10^\circ 5{\rm{cm}}$ settings. The $IoU_{3D}$ metric quantifies the overlap between the estimated and ground-truth 3D bounding boxes and is primarily used to assess size estimation accuracy. A prediction is considered correct when $IoU_{3D}$ exceeds a threshold of 75$\%$. In addition to accuracy, we assess model deployability and efficiency by reporting the GPU memory consumption (GB) during inference and running speed (FPS).

\subsection{Comparisons with State-of-the-Art Methods}
After training on the synthetic CAMERA25 dataset \cite{NOCS}, we first conduct quantitative comparisons between GenCOPE and state-of-the-art (SOTA) methods on the publicly available real-world REAL275 \cite{NOCS} and the in-the-wild Wild6D \cite{Wild6D} benchmarks. Then, we further perform qualitative comparisons under the same settings and finally evaluate domain generalization performance in real-world robotic manipulation scenarios.

\noindent\textbf{Quantitative Analysis on REAL275 and Wild6D} The results are reported in Table \ref{table1} and Table \ref{table2}. These two datasets represent complementary real-world challenges: REAL275 focuses on controlled real-world evaluation, while Wild6D emphasizes severe domain shift and unconstrained object appearances. For a fair comparison, for baselines with released models trained on CAMERA25, we directly evaluate them on real-world benchmarks; otherwise, we retrain using their official code on CAMERA25 only, without changing any settings, and evaluate on real-world benchmarks.

\begin{table*}[t!]
\renewcommand{\arraystretch}{1.1}
    \centering
    \tiny
    \caption{Comparison on the REAL275 dataset in metrics of $Io{U_{3D}}$ ($\%$) and $n^\circ m{\rm{cm}}$ ($\%$). "\checkmark" and "-" indicate with and without. "Syn" and "Real w/o Label" indicate synthetic dataset and real-world dataset without label. We use the $Io{U_{3D}}$ metric of CATRE \cite{CATRE}, which corrects NOCS \cite{NOCS} for size evaluation. The best results are bolded.}
    \vspace{1em}
    \resizebox{\textwidth}{!}{
    \begin{tabular}{l|ccc|c|cccc}
    \toprule[1.25pt]
    \multirow{2}[2]{*}{Method} & \multirow{2}[2]{*}{\begin{tabular}[c]{@{}c@{}} Syn \end{tabular}} & \multirow{2}[2]{*}{\begin{tabular}[c]{@{}c@{}} Real \\ w/o Label \end{tabular}} & \multirow{2}[2]{*}{\begin{tabular}[c]{@{}c@{}} Shape \\ Prior \end{tabular}} & \multicolumn{5}{c}{mean Average Precision (mAP)} \\
\cmidrule{5-9}       &       &    &      & ${\rm{3}}{{\rm{D}}_{75}}$ & $5^\circ$$2{\rm{cm}}$ & $5^\circ$$5{\rm{cm}}$ & $10^\circ$$2{\rm{cm}}$ & $10^\circ$$5{\rm{cm}}$ \\
    \midrule
    \multicolumn{9}{c}{Segmented by Mask R-CNN \cite{MaskR-CNN} Pretrained on ImageNet Dataset} \\
    \midrule
    SPD \cite{SPD}  & \checkmark       & -   & \checkmark    & 17.1  & 11.4  & 12.0  & 33.5  & 37.8  \\
    SGPA \cite{SGPA}  & \checkmark      & -  & \checkmark      & 17.8  & 19.8  & 27.7  & 36.5  & 62.6  \\
    STG6D \cite{STG6D} & \checkmark      & -  & \checkmark      & 14.4   & 20.1 & 27.9   & 39.0     & 63.7  \\
    SAR-Net \cite{SAR-Net} & \checkmark    & -  & \checkmark      & -   & 31.6 & 42.3   & 50.3     & 68.3  \\
    RePoNet \cite{Wild6D} & \checkmark     & \checkmark  & \checkmark     & -  & 29.1  & 31.3  & 48.5  & 56.8  \\
    SSC6D+ICP \cite{SSC6D} & \checkmark     & \checkmark   & -     & - & 28.6  & 33.4  & 51.8  & 62.9  \\
    DiffusionNOCS \cite{diffusionnocs} & \checkmark & - & -  & - & - & 35.0 & - & 66.6 \\
    CleanPose \cite{cleanpose}  & \checkmark & - & -  & 39.9 & 39.5 & 49.6 & 51.0 & 66.2 \\
    Diff9D \cite{Diff9D} & \checkmark       & -   & -     & 41.7 & 35.3 & 43.9 & 54.8 & 70.0 \\
    GenCOPE (ours) & \checkmark & - & -   & \textbf{42.6} & \textbf{39.7} & \textbf{50.0} & \textbf{57.0} & \textbf{71.6} \\
    \midrule
    \multicolumn{9}{c}{Segmented by Mask R-CNN \cite{MaskR-CNN} Pretrained on CAMERA25 Dataset} \\
    \midrule
    DPDN \cite{DPDN} & \checkmark      & -  & \checkmark      & 39.8  & 29.7  & 37.3  & 53.7  & 67.0  \\
    UDA-COPE \cite{UDA-COPE} & \checkmark      & \checkmark   & -      & 34.4  & 30.5  & 34.9  & 57.0  & 66.1  \\
    TTA-COPE \cite{TTA-COPE} & \checkmark      & -   & -     & 39.7  & 30.2  & 35.9  & \textbf{61.7}  & 73.2  \\
    VI-Net \cite{VI-Net} & \checkmark      & -   & -     & 11.3 & 11.7 & 19.9 & 14.4 & 29.1 \\
    SecondPose \cite{SecondPose} & \checkmark    & -   & -     & 14.7 & 12.7 & 21.2 & 17.6 & 34.2 \\
    CleanPose \cite{cleanpose}  & \checkmark & - & -   & 42.6 & 37.9 & 48.7 & 51.8 & 67.8 \\
    Diff9D \cite{Diff9D} & \checkmark      & -   & -     & 44.1 & 36.5 & 45.2 & 57.7 & 72.2 \\
    GenCOPE (ours) & \checkmark & - & -   & \textbf{44.5} & \textbf{38.1} & \textbf{49.9} & 56.4 & \textbf{74.4} \\
    \bottomrule[1.25pt]
    \end{tabular}
    }
  \label{table1}
  \vspace{-1em}
\end{table*}

\noindent \emph{REAL275 Dataset:} 
To comprehensively evaluate the robustness of GenCOPE under different segmentation qualities, we conduct experiments using two segmentation protocols. First, we adopt Mask R-CNN \cite{MaskR-CNN} pretrained on ImageNet \cite{imagenet} to obtain object masks. The corresponding results are shown in the upper part of Table \ref{table1}. Under identical experimental settings, GenCOPE achieves the best performance across all evaluation metrics without relying on any shape priors, consistently outperforming self-supervised approaches such as RePoNet \cite{Wild6D} and SSC6D \cite{SSC6D}, as well as domain generalization methods including DiffusionNOCS \cite{diffusionnocs} and Diff9D \cite{Diff9D}. To further ensure that no real-world training data is used throughout the entire pipeline, we employ Mask R-CNN trained solely on the synthetic CAMERA25 dataset \cite{NOCS} for instance segmentation. The results under this setting are summarized in the lower part of Table \ref{table1}. GenCOPE again shows superior performance compared to label-efficient methods, including DPDN \cite{DPDN}, UDA-COPE \cite{UDA-COPE}, TTA-COPE \cite{TTA-COPE}, and Diff9D \cite{Diff9D}, confirming its robustness under more challenging segmentation conditions. For a fair comparison, all methods are trained exclusively on the synthetic CAMERA25 dataset. While some methods further leverage the unlabeled REAL275 dataset \cite{NOCS} during training, GenCOPE does not utilize any real-world data. Moreover, when comparing against domain-specific SOTA methods with publicly available implementations such as VI-Net \cite{VI-Net}, SecondPose \cite{SecondPose}, and CleanPose \cite{cleanpose}, we evaluate their models trained on CAMERA25 only, ensuring full consistency with the GenCOPE training protocol. Overall, GenCOPE outperforms all these methods on REAL275 across both $IoU_{3D}$ and $n^\circ m{\rm{cm}}$ metrics, indicating a more reliable holistic pose estimation under real-world sensors.

\begin{table*}[t!]
\renewcommand{\arraystretch}{1.1}
\tiny
    \centering
    \caption{Comparison on the Wild6D dataset. "Real" indicates training with the CAMERA25 and REAL275 datasets. "\checkmark", "-", and "Syn" represent the same meanings as explained in Table \ref{table1}.}
    \vspace{1em}
    \resizebox{\linewidth}{!}{
    \begin{tabular}{l|ccc|c|cccc}
    \toprule[1.25pt]
    \multirow{2}[2]{*}{Method}    & \multirow{2}[2]{*}{\begin{tabular}[c]{@{}c@{}} Syn \end{tabular}} & \multirow{2}[2]{*}{\begin{tabular}[c]{@{}c@{}} Real \end{tabular}} & \multirow{2}[2]{*}{\begin{tabular}[c]{@{}c@{}} Shape \\ Prior \end{tabular}}  & \multicolumn{5}{c}{mean Average Precision (mAP)} \\
\cmidrule{5-9}          &       &       &     & ${\rm{3}}{{\rm{D}}_{75}}$ & $5^\circ$$2{\rm{cm}}$ & $5^\circ$$5{\rm{cm}}$ & $10^\circ$$2{\rm{cm}}$ & $10^\circ$$5{\rm{cm}}$ \\
    \midrule
    SPD \cite{SPD} & \checkmark      & \checkmark      & \checkmark      &  20.26     & 6.90   & 9.26   & 20.10   & 27.84  \\
    SGPA \cite{SGPA} & \checkmark      & \checkmark      & \checkmark       &  34.46   &  26.28   &  29.18   &  \textbf{33.80}   & 39.52  \\
    GPV-Pose \cite{GPV-Pose} & \checkmark      & \checkmark     & -       & -     & 14.10  & 21.50  & 23.80  & 41.10  \\
    Diff9D \cite{Diff9D} & \checkmark      & -     & -      &  38.16  & 25.52  & 30.46  & 32.48  & 40.94   \\
    GenCOPE (ours) & \checkmark      & -     & -       &  \textbf{38.72}     &   \textbf{26.66}    &   \textbf{33.32}    &   33.12    & \textbf{42.74}  \\
    \bottomrule[1.25pt]
    \end{tabular}
    }
  \label{table2}
\end{table*}

\begin{figure*}[t!]
\vspace{-1em}
\centering
\includegraphics[width=0.98\linewidth]{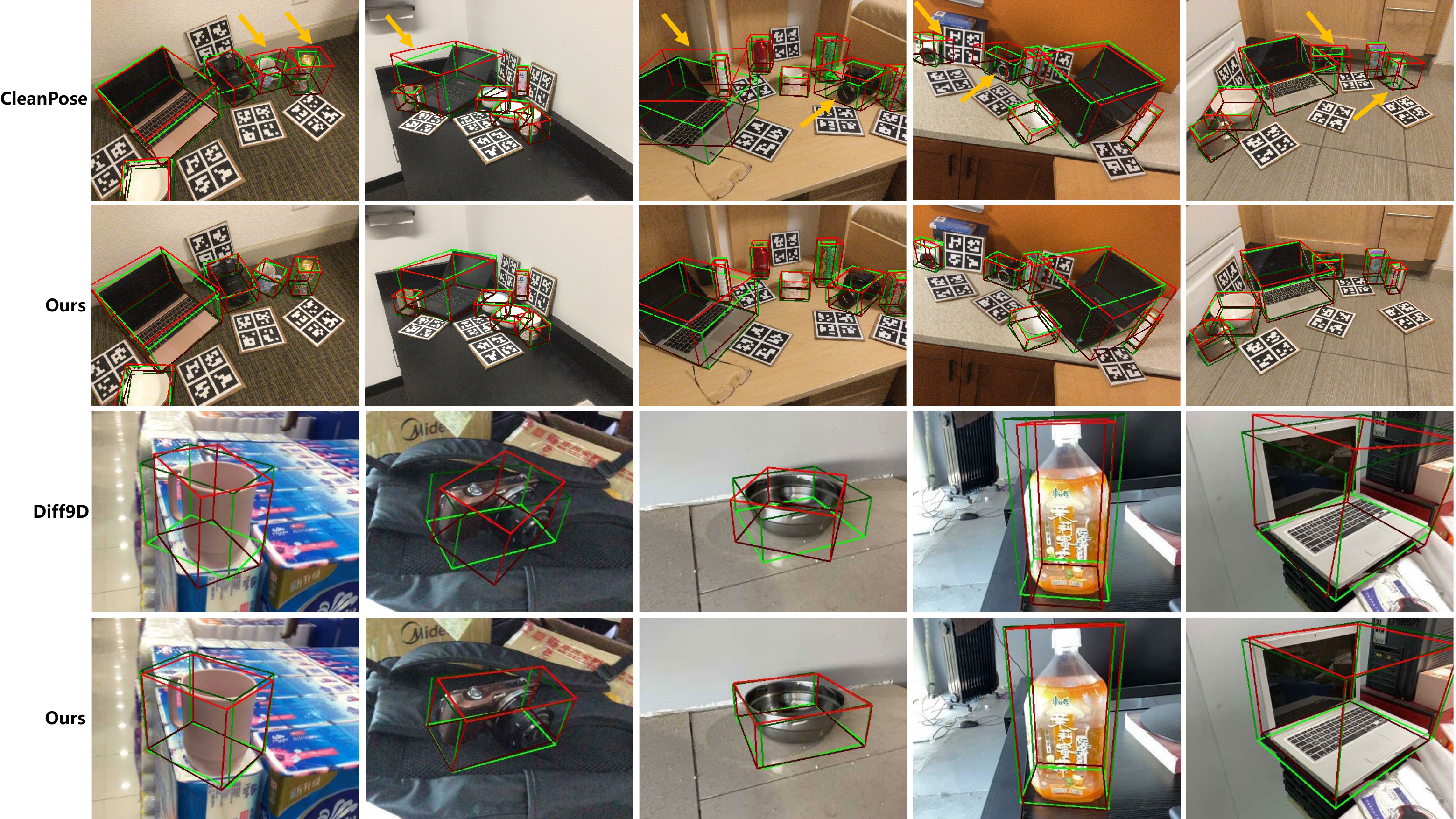}
\caption{Qualitative results comparison with CleanPose \cite{cleanpose} on REAL275 and Diff9D \cite{Diff9D} on Wild6D. The red and green boxes denote the estimated and ground-truth results, respectively.}
\label{Fig3}
\vspace{-1em}
\end{figure*}

\noindent \emph{Wild6D Dataset:}
The quantitative results on Wild6D dataset \cite{Wild6D}, as illustrated in Table \ref{table2}, further validate the robustness of GenCOPE under extreme domain shifts and in-the-wild conditions. The Wild6D dataset contains cluttered scenes, diverse lighting conditions, and background interference, posing challenges to existing methods. GenCOPE achieves the best overall performance in terms of both $IoU_{3D}$ and $n^\circ m{\rm{cm}}$ metrics, surpassing fully supervised approaches SPD \cite{SPD}, SGPA \cite{SGPA}, and GPV-Pose \cite{GPV-Pose}, as well as Diff9D \cite{Diff9D} under the same domain-generalized experimental setting. This proves that the cross-modal semantic consistency supervision suppresses over-dependence on domain-specific information and encourages semantic-aware feature learning. Consequently, GenCOPE maintains stable accuracy even when object appearances deviate from the training distribution.

\begin{wraptable}{r}{0.5\columnwidth}
\renewcommand{\arraystretch}{1.25}
    \centering
    \vspace{-1.5em}
    \caption{Analysis of model deployability and efficiency on the same computing platform. Infer. Memory indicates GPU memory usage at inference, and Speed measures the inference speed.}
    \vspace{0.5em}
    \resizebox{\linewidth}{!}{
    \begin{tabular}{l|c|c}
    \toprule[1.5pt]
       Method & Infer. Memory (GB) $\downarrow$ & Speed (FPS) $\uparrow$ \\
    \hline
       CleanPose \cite{cleanpose} & 10.04   & 24.33 \\
       Diff9D \cite{Diff9D} & 9.60  & 17.20 \\
       GenCOPE (ours) & \textbf{2.89}  & \textbf{40.68} \\ 
    \bottomrule[1.5pt]
    \end{tabular}
    }
    \label{table3}
    \vspace{-0.5em}
\end{wraptable}
\noindent \emph{Deployability and Efficiency Analysis:}
To enable practical deployment on real-world robotic platforms, model deployability and runtime efficiency are essential considerations. We therefore compare GenCOPE with CleanPose \cite{cleanpose} and Diff9D \cite{Diff9D} on the same computing platform, in terms of GPU memory consumption at inference and inference speed, as summarized in Table \ref{table3}. GenCOPE requires the least computational resources while achieving the fastest inference speed. This efficiency stems from the use of lightweight 2D and 3D encoders, the exclusive reliance on global features during 2D-3D cross consistency learning, and a compact pose-decoupled decoder. As a result, GenCOPE exhibits strong deployability and high efficiency, making it well-suited for real-world robotic platform, as mentioned in Implementation Details.

\begin{figure*}[t!]
\centering
\includegraphics[width=0.96\linewidth]{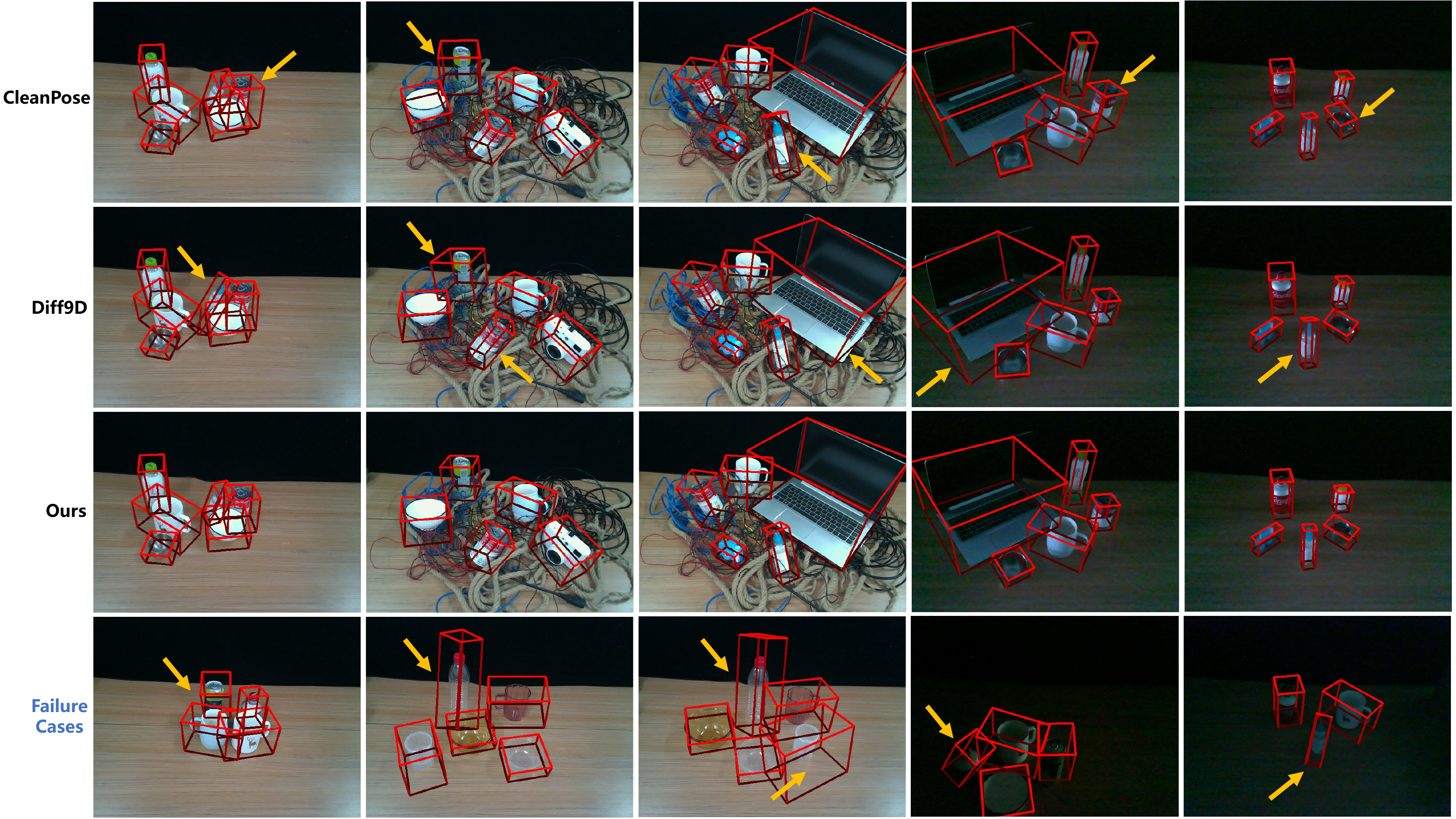}
\vspace{-0.5em}
\caption{Qualitative comparison with CleanPose \cite{cleanpose} and Diff9D \cite{Diff9D} in real-world robotic scenarios.}
\label{Fig4}
\vspace{-1em}
\end{figure*}

\noindent\textbf{Qualitative Analysis}
We deploy the trained, lightweight GenCOPE model on our robotic platform (with only 6 GB GPU memory) for on-board inference. In contrast, due to the higher GPU memory requirements of CleanPose \cite{cleanpose} and Diff9D \cite{Diff9D}, the captured RGB-D images are transmitted to the same workstation used during training for inference. For a fair comparison, all methods rely on identical object segmentation results obtained from Mask R-CNN \cite{MaskR-CNN}.

\noindent \emph{Qualitative Analysis on REAL275 and Wild6D Datasets:}
Figure \ref{Fig3} presents qualitative comparisons between GenCOPE with CleanPose \cite{cleanpose} and Diff9D \cite{Diff9D} on the REAL275 dataset \cite{NOCS} and Wild6D dataset \cite{Wild6D}, respectively. On the REAL275 dataset (first two rows), compared with CleanPose, GenCOPE produces more accurate 3D bounding boxes under cluttered scenes. In particular, CleanPose suffers from noticeable rotation and size deviations, as highlighted by the yellow arrows, whereas GenCOPE yields tighter alignment with ground-truth poses. On the Wild6D dataset (last two rows), which features diverse real-world scenes with complex backgrounds and severe domain shifts, GenCOPE consistently outperforms Diff9D. While Diff9D occasionally exhibits misaligned orientations, GenCOPE maintains stable pose predictions across different object categories. These results show that explicitly enforcing semantic consistency across modalities enables GenCOPE to suppress domain-specific cues better and achieve stronger domain generalization.

\noindent \emph{Qualitative Analysis in Real World:}
Figure \ref{Fig4} further evaluates GenCOPE in real-world robotic scenarios, where multiple objects are placed in occluded, cluttered, and low-light environments. Compared with CleanPose \cite{cleanpose} and Diff9D \cite{Diff9D}, GenCOPE produces more accurate pose estimations. These results show that the compared methods tend to produce incorrect orientations in complex scenes, while GenCOPE delivers more reliable predictions. In addition, the last row of Fig. \ref{Fig4} illustrates some representative failure cases, which mainly arise under severe occlusion, transparent objects with incomplete depth observations, or extremely low-light conditions (these scenarios still have significant domain gaps compared to the training data). Despite some limitations, GenCOPE demonstrates improved robustness compared to existing methods. Overall, these real-world results validate the practical effectiveness and deployability of GenCOPE for robotic tasks, aligning well with the quantitative evaluations reported earlier. \textbf{Some robotic picking demos} are provided in the Appendix \ref{GenCOPE Guided Robotic Picking} and the supplementary video.

\vspace{-0.5em}
\subsection{Ablation Studies}
\vspace{-0.5em}
We conduct ablations on two key components to quantify their contributions to GenCOPE on REAL275 dataset \cite{NOCS} (Table \ref{table4}): (i) the 2D/3D semantic consistency encoders, the loss of semantic consistency supervision, and the choice of LLM (Rows~A); and (ii) the 2D-3D cross consistency learning module and the self-/cross-attention mechanisms within the 2D-3D cross transformer (Rows~B). \textbf{More ablation studies} on the use of text feature ${F}_{text}$ with or without 2D/3D semantic consistency supervision are provided in the Appendix \ref{Additional Ablation Studies}.

\begin{wraptable}{r}{0.6\columnwidth}
\renewcommand{\arraystretch}{1.2}
    \centering
    \vspace{-1.5em}
    \caption{Ablation studies for the 2D/3D semantic consistency encoders and 2D-3D cross consistency learning module on the REAL275 dataset \cite{NOCS}.}
    \vspace{0.5em}
    \resizebox{\linewidth}{!}{
    \begin{tabular}{l|l|c|cccc}
    \toprule[1.5pt]
    \multirow{2}[2]{*}{Row} & \multirow{2}[2]{*}{Module Removal} & \multicolumn{5}{c}{mean Average Precision (mAP)} \\
\cmidrule{3-7}    & & ${\rm{3}}{{\rm{D}}_{75}}$ & $5^\circ$$2{\rm{cm}}$ & $5^\circ$$5{\rm{cm}}$ & $10^\circ$$2{\rm{cm}}$ & $10^\circ$$5{\rm{cm}}$ \\
    \midrule
    A0 & 2D Semantic Consistency    & 42.6  & 34.0  & 46.2  & 52.8  & 72.1 \\
    A1 & 3D Semantic Consistency   & 43.1  & 36.5  & 47.2  & 54.6  & 73.3 \\
    A2 & Class-Name-Only Text   & 41.9  & 33.4  & 45.8  & 52.1  & 71.7 \\
    A3 & $\mathcal{L}_{i}$ $ \to $ $\mathcal{L}_{cos}$   & 41.7  & 33.1  & 45.0  & 51.6  & 70.9\\
    A4 & Gemini-3 \cite{team2023gemini} $ \to $ Qwen2.5-VL-7B \cite{Qwen2.5-VL}     & 43.9  & 37.7  & 49.4  & 56.8  & 74.3 \\
    A5 & Gemini-3 \cite{team2023gemini} $ \to $ GPT-4 \cite{achiam2023gpt}     & 44.3   & 38.0 & 49.6   & \textbf{56.9}  & \textbf{74.6} \\
    \midrule
    B0 & 2D-3D Cross Consistency    & 42.9   & 34.9 & 47.0   & 53.8  & 72.9 \\
    B1 & $\mathrm{SelfAttn}(\cdot)$ in (e) of Fig. \ref{Fig2}     & 44.1  & 37.3 & 48.5 & 55.6  & 73.6 \\
    B2 & $\mathrm{CrossAttn}(\cdot)$ in (e) of Fig. \ref{Fig2}     & 43.3 & 36.0 & 47.3 & 54.7  & 73.1 \\
    \midrule
    C0 & Original GenCOPE    & \textbf{44.5}   & \textbf{38.1} & \textbf{49.9}   & 56.4  & 74.4 \\
    \bottomrule[1.5pt]
    \end{tabular}
    }
  \label{table4}
\vspace{-0.5em}
\end{wraptable}

\noindent\textbf{2D/3D Semantic Consistency Encoders}
First, we remove semantic consistency supervision from either encoder, which consistently degrades performance (Table \ref{table4} A0 and A1), confirming its necessity for robust representation learning. In particular, removing 2D consistency causes a larger regression (e.g., $38.1\%\!\rightarrow\!34.0\%$ at $5^\circ$$2{\rm{cm}}$) than removing the 3D counterpart ($38.1\%\!\rightarrow\!36.5$\%), suggesting that 2D features are more vulnerable to the Syn2Real domain gap and thus benefit more from explicit semantic alignment. Further, we conduct a finer-grained ablation by replacing part-structured descriptions with class-name-only text (e.g., "camera", "laptop", etc) for both 2D and 3D encoders, and observe a clear performance drop (Table \ref{table4} A2). This result confirms that the improvement arises from the structured semantic design rather than class-level supervision, providing direct evidence that intra-class semantic consistency alleviates the Syn2Real domain gap. Next, we replace our loss with a cosine similarity loss $\mathcal{L}_{cos}$ (Table \ref{table4} A3), which further incurs a substantial drop in all metrics, indicating that our formulation provides more effective supervision within a training batch containing multiple positive samples. Finally, we replace Gemini-3 \cite{team2023gemini} with Qwen2.5-VL-7B \cite{Qwen2.5-VL} and GPT-4 \cite{achiam2023gpt} while keeping the semantic query template and all other settings unchanged, generating comparable accuracy (Table \ref{table4} A4 and A5 vs.\ C0), which highlights that GenCOPE is robust to the choice of LLM.

\noindent\textbf{2D-3D Cross Consistency Learning}
When the module is completely removed and the 2D and 3D features are directly concatenated for pose regression (Table \ref{table4} B0), the accuracy drops across all metrics, validating the importance of cross-modal fusion. We then separately ablate the self-/cross-attention components within the 2D-3D cross transformer (Table \ref{table4} B1 and B2), both of which result in degradation. These results validate the necessity of the proposed module design and demonstrate that explicitly modeling cross-modal interactions between 2D and 3D features is crucial for accurate pose estimation.

\vspace{-0.5em}
\section{Conclusion} \label{Conclusion}
\vspace{-0.5em}
We presented GenCOPE, a simple yet effective Syn2Real generalized COPE method that enables models trained solely on synthetic data to generalize robustly to real-world deployment. We first analyze the root causes of domain discrepancies between synthetic and real-world data, and guided by these findings, introduced semantic consistency-guided 2D and 3D encoders and the 2D-3D cross consistency learning network with an end-to-end pose-decoupled decoder. Extensive results show that GenCOPE achieves superior domain generalization with minimal computational overhead and real-time inference, making it well-suited for practical integration into real-world robotic platforms.

\noindent\textbf{Limitation}
Although GenCOPE achieves encouraging results, the current formulation does not explicitly account for pose equivalence in strongly symmetric categories, where multiple rotations can describe the same physical configuration. Incorporating symmetry-aware losses and evaluation protocols remains an important direction for future work.

\vspace{-0.5em}
\section*{Acknowledgments}
\vspace{-0.5em}
This work is supported by the National Natural Science Foundation of China under Grants 62473141 and U22A2059, and also supported by the National Science and Technology Major Project of China under Grant 2026ZD1610900.

\medskip
{
\small
\bibliographystyle{unsrt}
\bibliography{main}

@String(AAAI = {AAAI})

@inproceedings{pspnet,
  title={Pyramid scene parsing network},
  author={Zhao, Hengshuang and Shi, Jianping and Qi, Xiaojuan and Wang, Xiaogang and Jia, Jiaya},
  booktitle={Proceedings of the IEEE conference on computer vision and pattern recognition},
  pages={2881--2890},
  year={2017}
}

@ARTICLE{PVNet,
  author={Peng, Sida and Zhou, Xiaowei and Liu, Yuan and Lin, Haotong and Huang, Qixing and Bao, Hujun},
  journal={IEEE Transactions on Pattern Analysis and Machine Intelligence}, 
  title={PVNet: Pixel-Wise Voting Network for 6DoF Object Pose Estimation}, 
  year={2022},
  volume={44},
  number={6},
  pages={3212-3223}
}

@inproceedings{PVN3D,
  title={Pvn3d: A deep point-wise 3d keypoints voting network for 6dof pose estimation},
  author={He, Yisheng and Sun, Wei and Huang, Haibin and Liu, Jianran and Fan, Haoqiang and Sun, Jian},
  booktitle={Proceedings of the IEEE/CVF Conference on Computer Vision and Pattern Recognition},
  pages={11632--11641},
  year={2020}
}

@article{liu2026survey,
  title={Deep learning-based object pose estimation: a comprehensive survey},
  author={Liu, Jian and Sun, Wei and Yang, Hui and Zeng, Zhiwen and Liu, Chongpei and Zheng, Jin and Liu, Xingyu and Rahmani, Hossein and Sebe, Nicu and Mian, Ajmal},  
  journal={International Journal of Computer Vision},
  volume={134},
  number={2},
  pages={81},
  year={2026}
}

@inproceedings{DenseFusion,
  title={Densefusion: 6d object pose estimation by iterative dense fusion},
  author={Wang, Chen and Xu, Danfei and Zhu, Yuke and Mart{\'\i}n-Mart{\'\i}n, Roberto and Lu, Cewu and Fei-Fei, Li and Savarese, Silvio},
  booktitle={Proceedings of the IEEE/CVF Conference on Computer Vision and Pattern Recognition},
  pages={3343--3352},
  year={2019}
}

@inproceedings{FFB6D,
  title={Ffb6d: A full flow bidirectional fusion network for 6d pose estimation},
  author={He, Yisheng and Huang, Haibin and Fan, Haoqiang and Chen, Qifeng and Sun, Jian},
  booktitle={Proceedings of the IEEE/CVF Conference on Computer Vision and Pattern Recognition},
  pages={3003--3013},
  year={2021}
}

@inproceedings{GDR-Net,
  title={Gdr-net: Geometry-guided direct regression network for monocular 6d object pose estimation},
  author={Wang, Gu and Manhardt, Fabian and Tombari, Federico and Ji, Xiangyang},
  booktitle={Proceedings of the IEEE/CVF Conference on Computer Vision and Pattern Recognition},
  pages={16611--16621},
  year={2021}
}

@inproceedings{NOCS,
  title={Normalized object coordinate space for category-level 6d object pose and size estimation},
  author={Wang, He and Sridhar, Srinath and Huang, Jingwei and Valentin, Julien and Song, Shuran and Guibas, Leonidas J},
  booktitle={Proceedings of the IEEE/CVF Conference on Computer Vision and Pattern Recognition},
  pages={2642--2651},
  year={2019}
}

@inproceedings{SPD,
  title={Shape prior deformation for categorical 6d object pose and size estimation},
  author={Tian, Meng and Ang, Marcelo H and Lee, Gim Hee},
  booktitle={European Conference on Computer Vision},
  pages={530--546},
  year={2020}
}

@inproceedings{CR-Net,
  title={Category-level 6D object pose estimation via cascaded relation and recurrent reconstruction networks},
  author={Wang, Jiaze and Chen, Kai and Dou, Qi},
  booktitle={Proceedings of the IEEE/RSJ International Conference on Intelligent Robots and Systems},
  pages={4807--4814},
  year={2021}
}

@inproceedings{SGPA,
  title={Sgpa: Structure-guided prior adaptation for category-level 6d object pose estimation},
  author={Chen, Kai and Dou, Qi},
  booktitle={Proceedings of the IEEE/CVF International Conference on Computer Vision},
  pages={2773--2782},
  year={2021}
}

@article{STG6D,
  title={Robotic continuous grasping system by shape transformer-guided multi-object category-level 6-d pose estimation},
  author={Liu, Jian and Sun, Wei and Liu, Chongpei and Zhang, Xing and Fu, Qiang},
  journal={IEEE Transactions on Industrial Informatics},
  volume={19},
  number={11},
  pages={11171-11181},
  year={2023}
}

@inproceedings{RBP-Pose,
  title={RBP-Pose: Residual bounding box projection for category-level pose estimation},
  author={Zhang, Ruida and Di, Yan and Lou, Zhiqiang and Manhardt, Fabian and Tombari, Federico and Ji, Xiangyang},
  booktitle={European Conference on Computer Vision},
  pages={655--672},
  year={2022}
}

@inproceedings{SSP-Pose,
  title={SSP-Pose: Symmetry-Aware Shape Prior Deformation for Direct Category-Level Object Pose Estimation},
  author={Zhang, Ruida and Di, Yan and Manhardt, Fabian and Tombari, Federico and Ji, Xiangyang},
  booktitle={Proceedings of the IEEE/RSJ International Conference on Intelligent Robots and Systems},
  pages={7452--7459},
  year={2022}
}

@inproceedings{CATRE,
  title={Catre: Iterative point clouds alignment for category-level object pose refinement},
  author={Liu, Xingyu and Wang, Gu and Li, Yi and Ji, Xiangyang},
  booktitle={European Conference on Computer Vision},
  pages={499--516},
  year={2022}
}

@inproceedings{DualPoseNet,
  title={Dualposenet: Category-level 6d object pose and size estimation using dual pose network with refined learning of pose consistency},
  author={Lin, Jiehong and Wei, Zewei and Li, Zhihao and Xu, Songcen and Jia, Kui and Li, Yuanqing},
  booktitle={Proceedings of the IEEE/CVF International Conference on Computer Vision},
  pages={3560--3569},
  year={2021}
}

@inproceedings{FS-Net,
  title={Fs-net: Fast shape-based network for category-level 6d object pose estimation with decoupled rotation mechanism},
  author={Chen, Wei and Jia, Xi and Chang, Hyung Jin and Duan, Jinming and Shen, Linlin and Leonardis, Ales},
  booktitle={Proceedings of the IEEE/CVF Conference on Computer Vision and Pattern Recognition},
  pages={1581--1590},
  year={2021}
}

@inproceedings{GPV-Pose,
  title={Gpv-pose: Category-level object pose estimation via geometry-guided point-wise voting},
  author={Di, Yan and Zhang, Ruida and Lou, Zhiqiang and Manhardt, Fabian and Ji, Xiangyang and Navab, Nassir and Tombari, Federico},
  booktitle={Proceedings of the IEEE/CVF Conference on Computer Vision and Pattern Recognition},
  pages={6781--6791},
  year={2022}
}

@inproceedings{IST-Net,
  title={IST-Net: Prior-free Category-level Pose Estimation with Implicit Space Transformation},
  author={Liu, Jianhui and Chen, Yukang and Ye, Xiaoqing and Qi, Xiaojuan},
  booktitle={Proceedings of the IEEE/CVF International Conference on Computer Vision},
  pages={13978--13988},
  year={2023}
}

@inproceedings{VI-Net,
  title={Vi-net: Boosting category-level 6d object pose estimation via learning decoupled rotations on the spherical representations},
  author={Lin, Jiehong and Wei, Zewei and Zhang, Yabin and Jia, Kui},
  booktitle={Proceedings of the IEEE/CVF International Conference on Computer Vision},
  pages={14001--14011},
  year={2023}
}

@inproceedings{CPPF,
  title={Cppf: Towards robust category-level 9d pose estimation in the wild},
  author={You, Yang and Shi, Ruoxi and Wang, Weiming and Lu, Cewu},
  booktitle={Proceedings of the IEEE/CVF Conference on Computer Vision and Pattern Recognition},
  pages={6866--6875},
  year={2022}
}

@inproceedings{SAR-Net,
  title={Sar-net: Shape alignment and recovery network for category-level 6d object pose and size estimation},
  author={Lin, Haitao and Liu, Zichang and Cheang, Chilam and Fu, Yanwei and Guo, Guodong and Xue, Xiangyang},
  booktitle={Proceedings of the IEEE/CVF Conference on Computer Vision and Pattern Recognition},
  pages={6707--6717},
  year={2022}
}

@inproceedings{SSC6D,
  title={Self-supervised category-level 6D object pose estimation with deep implicit shape representation},
  author={Peng, Wanli and Yan, Jianhang and Wen, Hongtao and Sun, Yi},
  booktitle={Proceedings of the AAAI Conference on Artificial Intelligence},
  volume={36},
  number={2},
  pages={2082--2090},
  year={2022}
}

@inproceedings{UDA-COPE,
  title={UDA-COPE: unsupervised domain adaptation for category-level object pose estimation},
  author={Lee, Taeyeop and Lee, Byeong-Uk and Shin, Inkyu and Choe, Jaesung and Shin, Ukcheol and Kweon, In So and Yoon, Kuk-Jin},
  booktitle={Proceedings of the IEEE/CVF Conference on Computer Vision and Pattern Recognition},
  pages={14891--14900},
  year={2022}
}

@inproceedings{Wild6D,
  title={Category-level 6d object pose estimation in the wild: A semi-supervised learning approach and a new dataset},
  author={Ze, Yanjie and Wang, Xiaolong},
  booktitle={Advances in Neural Information Processing Systems},
  volume={35},
  pages={27469--27483},
  year={2022}
}

@inproceedings{TTA-COPE,
  title={TTA-COPE: Test-Time Adaptation for Category-Level Object Pose Estimation},
  author={Lee, Taeyeop and Tremblay, Jonathan and Blukis, Valts and Wen, Bowen and Lee, Byeong-Uk and Shin, Inkyu and Birchfield, Stan and Kweon, In So and Yoon, Kuk-Jin},
  booktitle={Proceedings of the IEEE/CVF Conference on Computer Vision and Pattern Recognition},
  pages={21285--21295},
  year={2023}
}

@inproceedings{GenPose,
  title={Generative Category-level Object Pose Estimation via Diffusion Models},
  author={Zhang, Jiyao and Wu, Mingdong and Dong, Hao},
  booktitle={Advances in Neural Information Processing Systems},
  volume={36},
  pages={1--14},
  year={2023}
}

@inproceedings{ResNet,
  title={Deep residual learning for image recognition},
  author={He, Kaiming and Zhang, Xiangyu and Ren, Shaoqing and Sun, Jian},
  booktitle={Proceedings of the IEEE/CVF Conference on Computer Vision and Pattern Recognition},
  pages={770--778},
  year={2016}
}

@inproceedings{PointNet,
  title={Pointnet: Deep learning on point sets for 3d classification and segmentation},
  author={Qi, Charles R and Su, Hao and Mo, Kaichun and Guibas, Leonidas J},
  booktitle={Proceedings of the IEEE/CVF Conference on Computer Vision and Pattern Recognition},
  pages={652--660},
  year={2017}
}

@inproceedings{ViT,
  title={An Image is Worth 16x16 Words: Transformers for Image Recognition at Scale},
  author={Dosovitskiy, Alexey and Beyer, Lucas and Kolesnikov, Alexander and Weissenborn, Dirk and Zhai, Xiaohua and Unterthiner, Thomas and Dehghani, Mostafa and Minderer, Matthias and Heigold, Georg and Gelly, Sylvain and others},
  booktitle={International Conference on Learning Representations},
  year={2021}
}

@inproceedings{MaskR-CNN,
  title={Mask r-cnn},
  author={He, Kaiming and Gkioxari, Georgia and Doll{\'a}r, Piotr and Girshick, Ross},
  booktitle={Proceedings of the IEEE International Conference on Computer Vision},
  pages={2961--2969},
  year={2017}
}

@inproceedings{CyclicLR,
  title={Cyclical learning rates for training neural networks},
  author={Smith, Leslie N},
  booktitle={2017 IEEE Winter Conference on Applications of Computer Vision},
  pages={464--472},
  year={2017}
}

@inproceedings{Synthesis,
  title={Category level object pose estimation via neural analysis-by-synthesis},
  author={Chen, Xu and Dong, Zijian and Song, Jie and Geiger, Andreas and Hilliges, Otmar},
  booktitle={European Conference on Computer Vision},
  pages={139--156},
  year={2020}
}

@inproceedings{OLD-Net,
  title={Object level depth reconstruction for category level 6d object pose estimation from monocular rgb image},
  author={Fan, Zhaoxin and Song, Zhenbo and Xu, Jian and Wang, Zhicheng and Wu, Kejian and Liu, Hongyan and He, Jun},
  booktitle={European Conference on Computer Vision},
  pages={220--236},
  year={2022}
}

@article{DINOv2,
  title={Dinov2: Learning robust visual features without supervision},
  author={Oquab, Maxime and Darcet, Timoth{\'e}e and Moutakanni, Th{\'e}o and Vo, Huy and Szafraniec, Marc and Khalidov, Vasil and Fernandez, Pierre and Haziza, Daniel and Massa, Francisco and El-Nouby, Alaaeldin and others},
  journal={arXiv preprint arXiv:2304.07193},
  year={2023}
}

@inproceedings{SecondPose,
  title={SecondPose: SE (3)-Consistent Dual-Stream Feature Fusion for Category-Level Pose Estimation},
  author={Chen, Yamei and Di, Yan and Zhai, Guangyao and Manhardt, Fabian and Zhang, Chenyangguang and Zhang, Ruida and Tombari, Federico and Navab, Nassir and Busam, Benjamin},
  booktitle={Proceedings of the IEEE/CVF Conference on Computer Vision and Pattern Recognition},
  pages={9959--9969},
  year={2024}
}

@inproceedings{BundleTrack,
  title={Bundletrack: 6d pose tracking for novel objects without instance or category-level 3d models},
  author={Wen, Bowen and Bekris, Kostas},
  booktitle={IEEE/RSJ International Conference on Intelligent Robots and Systems},
  pages={8067--8074},
  year={2021}
}

@InProceedings{PhoCaL,
    author={Wang, Pengyuan and Jung, HyunJun and Li, Yitong and Shen, Siyuan and Srikanth, Rahul Parthasarathy and Garattoni, Lorenzo and Meier, Sven and Navab, Nassir and Busam, Benjamin},
    title={Phocal: A multi-modal dataset for category-level object pose estimation with photometrically challenging objects},
    booktitle={Proceedings of the IEEE/CVF Conference on Computer Vision and Pattern Recognition},
    pages={21222-21231},
    year={2022}
}

@inproceedings{HouseCat6D,
  title={Housecat6d-a large-scale multi-modal category level 6d object perception dataset with household objects in realistic scenarios},
  author={Jung, HyunJun and Wu, Shun-Cheng and Ruhkamp, Patrick and Zhai, Guangyao and Schieber, Hannah and Rizzoli, Giulia and Wang, Pengyuan and Zhao, Hongcheng and Garattoni, Lorenzo and Meier, Sven and others},
  booktitle={Proceedings of the IEEE/CVF Conference on Computer Vision and Pattern Recognition},
  pages={22498--22508},
  year={2024}
}

@article{MH6D,
  title={Mh6d: Multi-hypothesis consistency learning for category-level 6-d object pose estimation},
  author={Liu, Jian and Sun, Wei and Liu, Chongpei and Yang, Hui and Zhang, Xing and Mian, Ajmal},
  journal={IEEE Transactions on Neural Networks and Learning Systems.},
  year={2025},
  volume={36},
  number={3},
  pages={4820-4833}
}

@inproceedings{GeoReF,
  title={Georef: Geometric alignment across shape variation for category-level object pose refinement},
  author={Zheng, Linfang and Tse, Tze Ho Elden and Wang, Chen and Sun, Yinghan and Chen, Hua and Leonardis, Ales and Zhang, Wei and Chang, Hyung Jin},
  booktitle={Proceedings of the IEEE/CVF Conference on Computer Vision and Pattern Recognition},
  pages={10693--10703},
  year={2024}
}

@inproceedings{AG-Pose,
  title={Instance-adaptive and geometric-aware keypoint learning for category-level 6d object pose estimation},
  author={Lin, Xiao and Yang, Wenfei and Gao, Yuan and Zhang, Tianzhu},
  booktitle={Proceedings of the IEEE/CVF Conference on Computer Vision and Pattern Recognition},
  pages={21040--21049},
  year={2024}
}

@inproceedings{DPDN,
  title={Category-level 6d object pose and size estimation using self-supervised deep prior deformation networks},
  author={Lin, Jiehong and Wei, Zewei and Ding, Changxing and Jia, Kui},
  booktitle={European Conference on Computer Vision},
  pages={19--34},
  year={2022}
}

@inproceedings{centersnap,
  title={Centersnap: Single-shot multi-object 3d shape reconstruction and categorical 6d pose and size estimation},
  author={Irshad, Muhammad Zubair and Kollar, Thomas and Laskey, Michael and Stone, Kevin and Kira, Zsolt},
  booktitle={International Conference on Robotics and Automation},
  pages={10632--10640},
  year={2022}
}

@inproceedings{diffusionnocs,
  title={Diffusionnocs: Managing symmetry and uncertainty in sim2real multi-modal category-level pose estimation},
  author={Ikeda, Takuya and Zakharov, Sergey and Ko, Tianyi and Irshad, Muhammad Zubair and Lee, Robert and Liu, Katherine and Ambrus, Rares and Nishiwaki, Koichi},
  booktitle={IEEE/RSJ International Conference on Intelligent Robots and Systems},
  pages={7406--7413},
  year={2024}
}

@article{Diff9D,
  title={Diff9d: Diffusion-based domain-generalized category-level 9-dof object pose estimation},
  author={Liu, Jian and Sun, Wei and Yang, Hui and Deng, Pengchao and Liu, Chongpei and Sebe, Nicu and Rahmani, Hossein and Mian, Ajmal},
  journal={IEEE Transactions on Pattern Analysis and Machine Intelligence},
  year={2025},
  volume={47},
  number={7},
  pages={5520-5537}
}

@inproceedings{cleanpose,
  title={Cleanpose: Category-level object pose estimation via causal learning and knowledge distillation},
  author={Lin, Xiao and Peng, Yun and Wang, Liuyi and Zhong, Xianyou and Zhu, Minghao and Feng, Yi and Yang, Jingwei and Liu, Chengju and Chen, Qijun},
  booktitle={Proceedings of the IEEE/CVF International Conference on Computer Vision},
  pages={5990--6000},
  year={2025}
}

@article{clipose,
  title={Clipose: Category-level object pose estimation with pre-trained vision-language knowledge},
  author={Lin, Xiao and Zhu, Minghao and Dang, Ronghao and Zhou, Guangliang and Shu, Shaolong and Lin, Feng and Liu, Chengju and Chen, Qijun},
  journal={arXiv preprint arXiv:2402.15726},
  year={2024}
}

@inproceedings{query6dof,
  title={Query6dof: Learning sparse queries as implicit shape prior for category-level 6dof pose estimation},
  author={Wang, Ruiqi and Wang, Xinggang and Li, Te and Yang, Rong and Wan, Minhong and Liu, Wenyu},
  booktitle={Proceedings of the IEEE/CVF International Conference on Computer Vision},
  pages={14055--14064},
  year={2023}
}

@inproceedings{GCE-Pose,
  title={Gce-pose: Global context enhancement for category-level object pose estimation},
  author={Li, Weihang and Xu, Hongli and Huang, Junwen and Jung, Hyunjun and Yu, Peter KT and Navab, Nassir and Busam, Benjamin},
  booktitle={Proceedings of the IEEE/CVF Conference on Computer Vision and Pattern Recognition},
  pages={27154--27165},
  year={2025}
}

@inproceedings{SpotPose,
  title={Rethinking correspondence-based category-level object pose estimation},
  author={Ren, Huan and Yang, Wenfei and Zhang, Shifeng and Zhang, Tianzhu},
  booktitle={Proceedings of the IEEE/CVF Conference on Computer Vision and Pattern Recognition},
  pages={1170--1179},
  year={2025}
}

@article{LanCOPE,
  title={Lancope: Language-guided category-level object pose estimation from a single rgb image},
  author={Yang, Hui and Sun, Wei and Liu, Jian and Zheng, Jin and Dai, Zhenqi and Mian, Ajmal},
  journal={IEEE Robotics and Automation Letters},
  year={2025}
}

@article{yang2025rgb,
  title={Rgb-based category-level object pose estimation via depth recovery and adaptive refinement},
  author={Yang, Hui and Sun, Wei and Liu, Jian and Zheng, Jin and Zeng, Zhiwen and Mian, Ajmal},
  journal={IEEE Robotics and Automation Letters},
  year={2025}
}

@inproceedings{clip,
  title={Learning transferable visual models from natural language supervision},
  author={Radford, Alec and Kim, Jong Wook and Hallacy, Chris and Ramesh, Aditya and Goh, Gabriel and Agarwal, Sandhini and Sastry, Girish and Askell, Amanda and Mishkin, Pamela and Clark, Jack and others},
  booktitle={International Conference on Machine Learning},
  pages={8748--8763},
  year={2021}
}

@inproceedings{khosla2020supervised,
  title={Supervised contrastive learning},
  author={Khosla, Prannay and Teterwak, Piotr and Wang, Chen and Sarna, Aaron and Tian, Yonglong and Isola, Phillip and Maschinot, Aaron and Liu, Ce and Krishnan, Dilip},
  booktitle={Advances in Neural Information Processing Systems},
  volume={33},
  pages={18661--18673},
  year={2020}
}

@article{team2023gemini,
  title={Gemini: A family of highly capable multimodal models},
  author={Team, Gemini and Anil, Rohan and Borgeaud, Sebastian and Alayrac, Jean-Baptiste and Yu, Jiahui and Soricut, Radu and Schalkwyk, Johan and Dai, Andrew M and Hauth, Anja and Millican, Katie and others},
  journal={arXiv preprint arXiv:2312.11805},
  year={2023}
}

@article{achiam2023gpt,
  title={Gpt-4 technical report},
  author={Achiam, Josh and Adler, Steven and Agarwal, Sandhini and Ahmad, Lama and Akkaya, Ilge and Aleman, Florencia Leoni and Almeida, Diogo and Altenschmidt, Janko and Altman, Sam and Anadkat, Shyamal and others},
  journal={arXiv preprint arXiv:2303.08774},
  year={2023}
}

@inproceedings{imagenet,
  title={Imagenet: A large-scale hierarchical image database},
  author={Deng, Jia and Dong, Wei and Socher, Richard and Li, Li-Jia and Li, Kai and Fei-Fei, Li},
  booktitle={Proceedings of the IEEE/CVF Conference on Computer Vision and Pattern Recognition},
  pages={248--255},
  year={2009}
}

@inproceedings{huang2025givepose,
  title={Givepose: Gradual intra-class variation elimination for rgb-based category-level object pose estimation},
  author={Huang, Ziqin and Wang, Gu and Zhang, Chenyangguang and Zhang, Ruida and Li, Xiu and Ji, Xiangyang},
  booktitle={Proceedings of the IEEE/CVF Conference on Computer Vision and Pattern Recognition},
  pages={22055--22066},
  year={2025}
}

@article{Qwen2.5-VL,
  title={Qwen2.5-VL Technical Report},
  author={Bai, Shuai and Chen, Keqin and Liu, Xuejing and Wang, Jialin and Ge, Wenbin and Song, Sibo and Dang, Kai and Wang, Peng and Wang, Shijie and Tang, Jun and Zhong, Humen and Zhu, Yuanzhi and Yang, Mingkun and Li, Zhaohai and Wan, Jianqiang and Wang, Pengfei and Ding, Wei and Fu, Zheren and Xu, Yiheng and Ye, Jiabo and Zhang, Xi and Xie, Tianbao and Cheng, Zesen and Zhang, Hang and Yang, Zhibo and Xu, Haiyang and Lin, Junyang},
  journal={arXiv preprint arXiv:2502.13923},
  year={2025}
}

@inproceedings{miao2023spectrum,
  title={Spectrum-guided multi-granularity referring video object segmentation},
  author={Miao, Bo and Bennamoun, Mohammed and Gao, Yongsheng and Mian, Ajmal},
  booktitle={IEEE/CVF International Conference on Computer Vision},
  pages={920--930},
  year={2023},
}

@article{miao2024referring,
  title={Referring human pose and mask estimation in the wild},
  author={Miao, Bo and Feng, Mingtao and Wu, Zijie and Bennamoun, Mohammed and Gao, Yongsheng and Mian, Ajmal},
  journal={Advances in Neural Information Processing Systems},
  volume={37},
  pages={44791--44813},
  year={2024}
}

@inproceedings{miao2026langmap,
  title={LangMap: A Human-Verified Benchmark for Hierarchical Open-Vocabulary Goal Navigation},
  author={Miao, Bo and Liu, Weijia and Luo, Jun and Shinnick, Lachlan and Liu, Jian and Hamilton-Smith, Thomas and Yang, Yuhe and Wu, Zijie and Videnovic, Vanja and Dayoub, Feras and van den Hengel, Anton},
  booktitle={Advances in Neural Information Processing Systems},
  year={2026}
}

@article{liu2026sinref6d,
  author={Liu, Jian and Sun, Wei and Zeng, Kai and Zheng, Jin and Yang, Hui and Rahmani, Hossein and Mian, Ajmal and Wang, Lin},
  journal={IEEE Transactions on Robotics}, 
  title={Scalable Unseen Objects 6-DoF Absolute Pose Estimation With Robotic Integration}, 
  year={2026},
  volume={42},
  pages={1884-1901}
}

@INPROCEEDINGS{monodiff9d,
  author={Liu, Jian and Sun, Wei and Yang, Hui and Zheng, Jin and Geng, Zichen and Rahmani, Hossein and Mian, Ajmal},
  booktitle={IEEE International Conference on Robotics and Automation}, 
  title={MonoDiff9D: Monocular Category-Level 9D Object Pose Estimation via Diffusion Model}, 
  year={2025},
  pages={8687-8694}
}

@INPROCEEDINGS{geng2026disentangled,
  title={Disentangled Hierarchical VAE for 3D Human-Human Interaction Generation},
  author={Geng, Zichen and Hayder, Zeeshan and Miao, Bo and Liu, Jian and Liu, Wei and Mian, Ajmal},
  booktitle={International Conference on Learning Representations},
  year={2026}
}

@article{yang2026occlusion,
  title={Occlusion-aware 3d hand-object pose estimation with masked autoencoders},
  author={Yang, Hui and Sun, Wei and Liu, Jian and Zheng, Jin and Xiao, Jian and Mian, Ajmal},
  journal={IEEE Transactions on Multimedia},
  year={2026}
}

@inproceedings{yang2026genhoi,
  title={GenHOI: Generalized Hand-Object Pose Estimation with Occlusion Awareness},
  author={Yang, Hui and Sun, Wei and Liu, Jian and Xiao, Jian and Xie, Tao and Rahmani, Hossein and Mian, Ajmal and Sebe, Nicu and Lee, Gim Hee},
  booktitle={European Conference on Computer Vision},
  pages={684--701},
  year={2026}
}

@article{yang2022deepinteraction,
  title={Deepinteraction: 3d object detection via modality interaction},
  author={Yang, Zeyu and Chen, Jiaqi and Miao, Zhenwei and Li, Wei and Zhu, Xiatian and Zhang, Li},
  journal={Advances in Neural Information Processing Systems},
  volume={35},
  pages={1992--2005},
  year={2022}
}

@inproceedings{zhao2021few,
  title={Few-shot 3d point cloud semantic segmentation},
  author={Zhao, Na and Chua, Tat-Seng and Lee, Gim Hee},
  booktitle={IEEE/CVF Conference on Computer Vision and Pattern Recognition},
  pages={8869--8878},
  year={2021}
}

@inproceedings{zhao2020sess,
  title={Sess: Self-ensembling semi-supervised 3d object detection},
  author={Zhao, Na and Chua, Tat-Seng and Lee, Gim Hee},
  booktitle={IEEE/CVF Conference on Computer Vision and Pattern Recognition},
  pages={11076--11084},
  year={2020}
}

@inproceedings{mkpose,
  title={Mk-pose: Category-level object pose estimation via multimodal-based keypoint learning},
  author={Yang, Yifan and Song, Peili and Lan, Enfan and Liu, Dong and Liu, Jingtai},
  booktitle={IEEE/RSJ International Conference on Intelligent Robots and Systems},
  pages={4232--4239},
  year={2025}
}

@inproceedings{lin2026mempose,
  title={MemPose: Category-Level Object Pose Estimation with Memory},
  author={Lin, Xiao and Zhu, Minghao and Peng, Yun and Wang, Liuyi and Wang, Qiyi and Liu, Chengju and Chen, Qijun},
  booktitle={European Conference on Computer Vision},
  pages={156--174},
  year={2026}
}

@inproceedings{ren2026compose,
  title={ComPose: A Unified Completion-Pose Framework for Robust Category-Level Object Pose Estimation},
  author={Ren, Huan and Chen, Yihan and Wang, Chuxin and Liu, Nailong and Yang, Wenfei and Zhang, Tianzhu},
  booktitle={Proceedings of the IEEE/CVF Conference on Computer Vision and Pattern Recognition},
  year={2026}
}

@inproceedings{zhang2024omni6dpose,
  title={Omni6dpose: A benchmark and model for universal 6d object pose estimation and tracking},
  author={Zhang, Jiyao and Huang, Weiyao and Peng, Bo and Wu, Mingdong and Hu, Fei and Chen, Zijian and Zhao, Bo and Dong, Hao},
  booktitle={European Conference on Computer Vision},
  pages={199--216},
  year={2024}
}
}

\newpage

\appendix
\section*{Appendix}

\section{Additional Semantic Samples Generated by LLM}\label{Additional Semantic Samples Generated by LLM}
\begin{figure*}[htbp]
\renewcommand{\thefigure}{A}
\centering
\includegraphics[width=\textwidth]{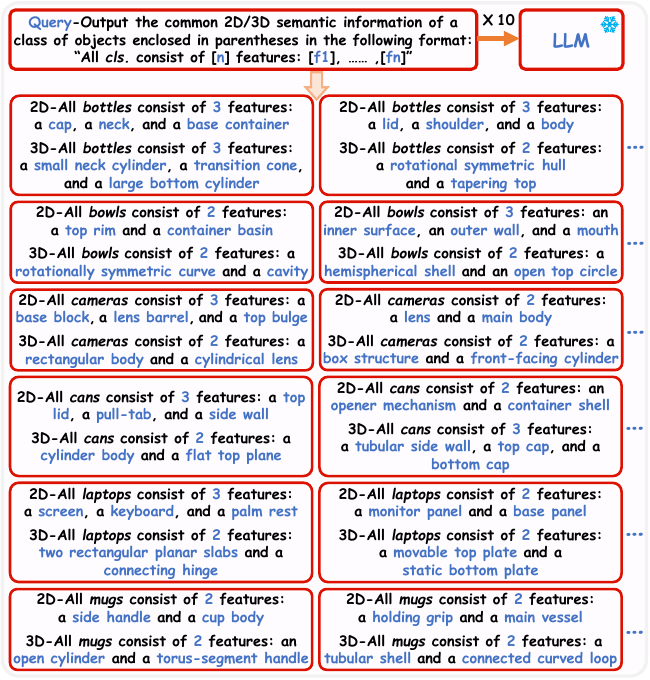}
\caption{Diverse semantic samples generated by LLM for different object categories.}
\label{SupFig1}
\end{figure*}

As illustrated in Fig. \ref{SupFig1}, the LLM \cite{team2023gemini} generates diverse semantic descriptions for different object categories. Even when queried with an identical template for the same category, the LLM produces varied responses across multiple calls. To exploit this variability, we query the LLM ten times per object category during training and randomly sample one description at each iteration, thereby increasing the diversity and stochasticity of semantic supervision.

\section{GenCOPE Guided Robotic Picking}\label{GenCOPE Guided Robotic Picking}
\begin{figure*}[htbp]
\renewcommand{\thefigure}{B}
\centering
\includegraphics[width=\textwidth]{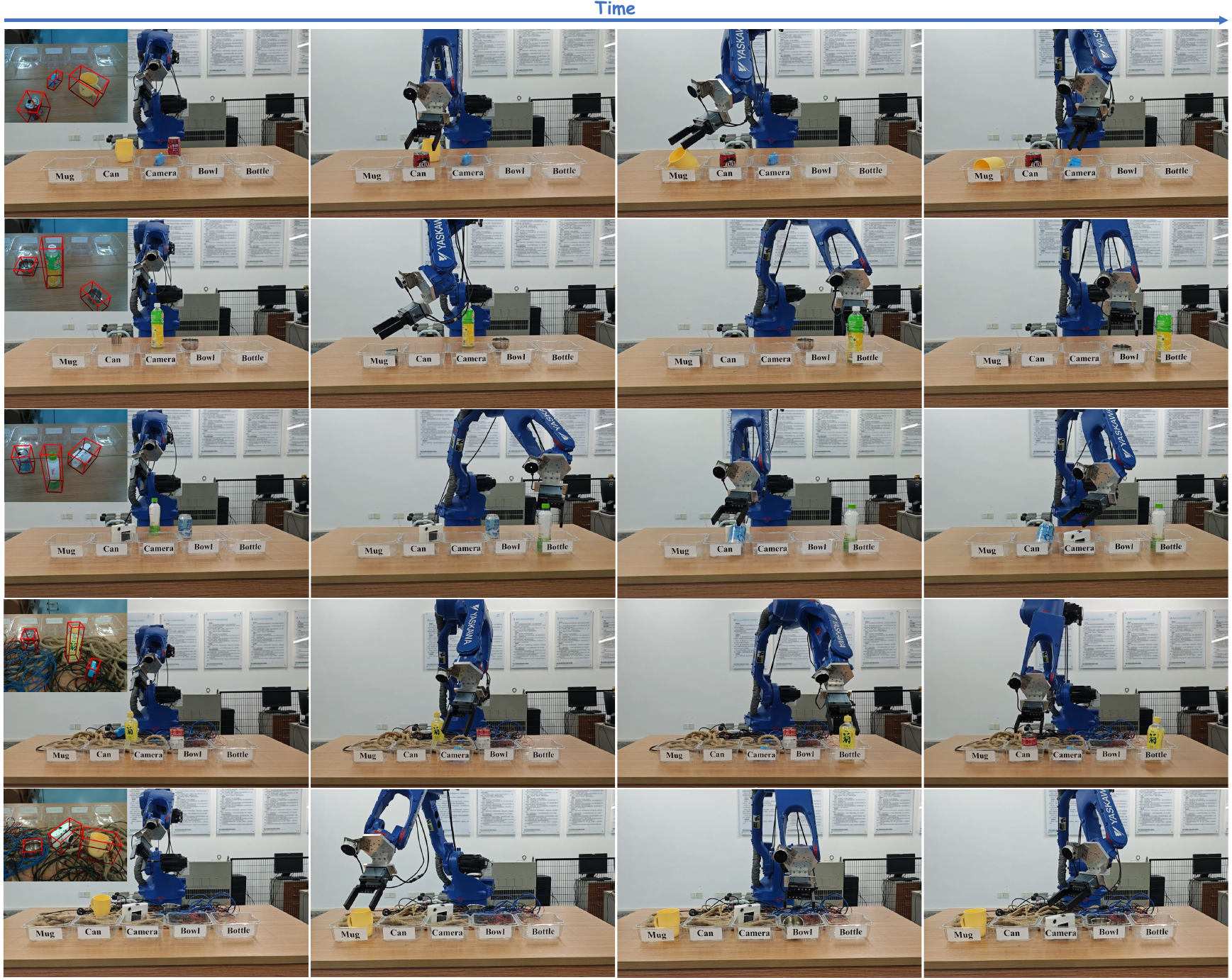}
\caption{Visualization of GenCOPE-guided multi-object continuous robotic picking.}
\label{SupFig2}
\end{figure*}

To validate the practicality of GenCOPE trained solely on synthetic data, we deploy it on a real-world eye-in-hand robotic picking platform equipped with an Intel RealSense L515 RGB-D camera and an onboard NVIDIA GTX 1660 GPU with 6 GB memory. Owing to its lightweight design, GenCOPE runs entirely on the embedded hardware without offloading computation. Representative pose estimation results are shown in the top-left of Fig. \ref{SupFig2}. To ensure safe grasp execution, the grasp position is manually offset by 2$cm$ above the tabletop, and the grasp orientation is tilted upward by 30°. Multiple objects in the scene, including textureless metallic components, are then sequentially picked based on their estimated center depths, as illustrated in Fig. \ref{SupFig2}. These real-world experiments demonstrate the strong deployability, high efficiency, and robust Syn2Real generalization of GenCOPE, confirming its effectiveness for robotic picking.

\section{Details of Some Modules}\label{Details of Some Modules}
\begin{figure*}[htbp]
\renewcommand{\thefigure}{C.1}
\centering
\includegraphics[width=\textwidth]{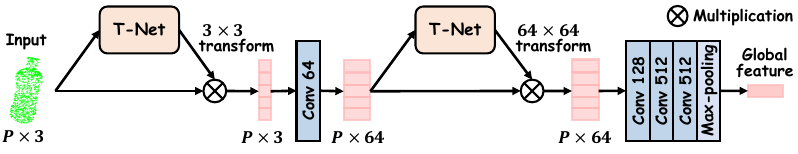}
\caption{Detailed structure of the PointNet-based \cite{PointNet} point cloud global feature encoder.}
\label{SupFig3}
\end{figure*}

\begin{figure*}[htbp]
\renewcommand{\thefigure}{C.2}
\centering
\includegraphics[width=0.65\textwidth]{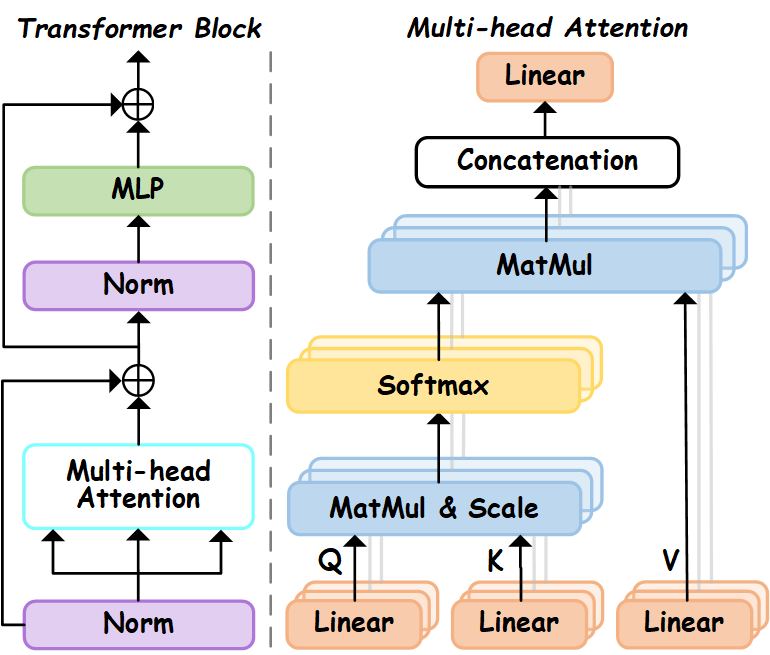}
\caption{Detailed structure of the transformer block \cite{ViT} used in self-attention and cross-attention modules. $ \oplus $ and "MatMul" represent residual connection and matrix multiplication, respectively.}
\label{SupFig4}
\end{figure*}

In the feature encoding stage, we adopt ResNet18 \cite{ResNet} with the final fully connected layer removed to extract 2D features from the input RGB image. For the point cloud back-projected from the depth map, we employ a PointNet-based encoder (detailed in Fig. \ref{SupFig3}); unlike the standard PointNet \cite{PointNet}, it is used exclusively for global feature extraction to prioritize deployability and computational efficiency. In addition, both the self-attention and cross-attention modules in the 2D-3D cross consistency learning network are implemented using a shared transformer block, as illustrated in Fig. \ref{SupFig4}.

\section{Additional Ablation Studies}\label{Additional Ablation Studies}

\begin{table*}[htbp]
\vspace{-1em}
\renewcommand\arraystretch{1.2}
\newcommand{\tabincell}[2]{\begin{tabular}{@{}#1@{}}#2\end{tabular}}
\renewcommand{\thetable}{A}
  \centering
  \setlength{\tabcolsep}{5pt} 
  \small
  \caption{Ablation studies on the use of text features $F_{\text{text}}$ (including $F_{\text{text-2D}}$ and $F_{\text{text-3D}}$), with and without 2D/3D semantic consistency supervision, on the REAL275 dataset \cite{NOCS}.}
  \vspace{0.5em}
    \begin{tabular}{l|cc|c|cccc}
    \toprule[1.5pt]
    \multirow{2}[2]{*}{Row} & \multirow{2}[2]{*}{\begin{tabular}[c]{@{}c@{}} ${F}_{text}$ in \\ Train. \& Infer. \end{tabular}} & \multirow{2}[2]{*}{\begin{tabular}[c]{@{}c@{}} $\mathcal{L}_{C\_text\text{-}2D}$ \\ \& $\mathcal{L}_{C\_text\text{-}3D}$ \end{tabular}}  & \multicolumn{5}{c}{mean Average Precision (mAP)} \\
\cmidrule{4-8}   & & & ${\rm{3}}{{\rm{D}}_{75}}$ & \tabincell{c}{$5^\circ$$2{\rm{cm}}$} & \tabincell{c}{$5^\circ$$5{\rm{cm}}$} & \tabincell{c}{$10^\circ$$2{\rm{cm}}$} & \tabincell{c}{$10^\circ$$5{\rm{cm}}$} \\
    \midrule
    S0 & residual & \textbf{--}   & 42.1  & 33.7  & 45.4  & 51.6  & 71.0 \\
    S1 & concatenate & \textbf{--}   & 42.5  & 34.6  & 46.1  & 52.5  & 72.1 \\
    \midrule
    S2 & residual  & \checkmark     & 44.0  & 38.0  & 49.6  & 55.9  & 74.0 \\
    S3 & concatenate   & \checkmark      & 44.4   & \textbf{38.3} & \textbf{50.0}   & 56.2  & \textbf{74.8} \\
    \midrule
    Ours & \textbf{--}   & \checkmark     & \textbf{44.5}   & 38.1 & 49.9   & \textbf{56.4}  & 74.4 \\
    \bottomrule[1.5pt]
    \end{tabular}
  \label{suptable1}
\end{table*}

Table \ref{suptable1} presents an ablation study analyzing the role of text features and the proposed 2D/3D semantic consistency supervision on the REAL275 dataset \cite{NOCS}. The second column evaluates whether CLIP-based \cite{clip} text features are introduced during both training and inference via direct residual addition or feature concatenation, while the third column examines the effect of enabling the proposed 2D and 3D semantic consistency supervision during training. Specifically, the direct residual addition is as follows:
\begin{equation}\label{supequation1}
\begin{aligned}
F_{C\text{-}2D} = F_{C\text{-}2D} + F_{text\text{-}2D},\\
F_{C\text{-}3D} = F_{C\text{-}3D} + F_{text\text{-}3D}.
\end{aligned}
\end{equation}
Then, $F_{C\text{-}2D}$ and $F_{C\text{-}3D}$ are fed into the same 2D-3D cross consistency learning module for pose regression. The feature concatenation is defined as follows:
\begin{equation}\label{supequation2}
\begin{aligned}
F_{C\text{-}2D} = \mathrm{Concat}\!\left(
F_{C\text{-}2D},\;
F_{text\text{-}2D}
\right),\\
F_{C\text{-}3D} = \mathrm{Concat}\!\left(
F_{C\text{-}3D},\;
F_{text\text{-}3D}
\right).
\end{aligned}
\end{equation}
Based on this, we double the feature dimensions of the 2D-3D cross consistency learning module and the pose decoder, while preserving the same network architecture, to align with the corresponding $F_{C\text{-}2D}$ and $F_{C\text{-}3D}$. The experimental results in Table \ref{suptable1} show that directly injecting text features into the visual encoders does not consistently improve performance (Table \ref{suptable1} S0 and S1). Feature concatenation achieves performance comparable to our full model (Table \ref{suptable1} S3), whereas residual addition slightly degrades (Table \ref{suptable1} S2).
Moreover, introducing text features during both training and inference inevitably increases computational cost and memory consumption, which is undesirable for real-time robotic deployment. In contrast, incorporating the proposed semantic consistency supervision consistently improves performance, regardless of whether text features are explicitly fused into the 2D/3D representations. These results indicate that enforcing semantic alignment during feature learning is more effective than directly multimodal feature fusion. Overall, the findings validate that the proposed 2D/3D semantic consistency supervision enables the encoders to focus on domain-invariant semantic cues, thereby improving robustness to domain-specific information while maintaining computational efficiency and training stability.


\end{document}